\documentclass[10pt,times,mathptm,psfig,twocolumn,journals]{IEEEtran}
\usepackage{url}
\usepackage{amssymb}
\usepackage{array}
\usepackage{amsmath}
\usepackage{graphicx}
\usepackage{booktabs}
\usepackage{epsfig}
\usepackage{ulem}
\usepackage{hyperref}
\hypersetup{pdfborder={0 0 0}, colorlinks=true, linkcolor=red, citecolor=blue}
\usepackage{subfigure}
\usepackage{cite}
\usepackage{epstopdf}
\usepackage{amsmath}
\usepackage[misc]{ifsym}
\usepackage{bm}
\usepackage{mathrsfs}
\usepackage{float}
\usepackage{setspace}
\usepackage{multirow}
\usepackage{tabularx}
\usepackage{url}
\usepackage{booktabs}
\usepackage{makecell}
\usepackage{epigraph}

\begin{document}

%\title{Unsupervised Deep Image Stitching: Reconstructing Stitched Images from Feature to Pixel}

%\uline{\title{Unsupervised Deep Image Stitching: Reconstructing Stitched Features to Images}}
\title{Unsupervised Deep Image Stitching: Reconstructing Stitched Features to Images}

\author{Lang~Nie, Chunyu~Lin, Kang~Liao, Shuaicheng Liu,~\IEEEmembership{Member,~IEEE}, Yao~Zhao,~\IEEEmembership{Senior Member,~IEEE}
\thanks{This work was supported by the National Natural Science
Foundation of China (No.61772066, No.61972028).
\textit{(Corresponding author: Chunyu Lin)}}
\thanks{Lang Nie, Chunyu Lin, Kang Liao, Yao Zhao are with the Institute of Information Science, Beijing Jiaotong University, Beijing 100044, China, and also with the Beijing Key Laboratory of Advanced Information Science and Network Technology, Beijing 100044, China (email: nielang@bjtu.edu.cn, cylin@bjtu.edu.cn, kang\_liao@bjtu.edu.cn, yzhao@bjtu.edu.cn).}
\thanks{Shuaicheng Liu is with School of Information and Communication Engineering, University of Electronic Science and Technology of China, Chengdu, 611731, China (liushuaicheng@uestc.edu.cn).}
}

\maketitle
%\graphicspath{{img/}}
\begin{abstract}
    Traditional feature-based image stitching technologies rely heavily on feature detection quality, often failing to stitch images with few features or low resolution. The learning-based image stitching solutions are rarely studied due to the lack of labeled data, making the supervised methods unreliable.
    To address the above limitations, we propose an unsupervised deep image stitching framework consisting of two stages: unsupervised coarse image alignment and unsupervised image reconstruction. In the first stage, we design an ablation-based loss to constrain an unsupervised homography network, which is more suitable for large-baseline scenes. Moreover, a transformer layer is introduced to warp the input images in the stitching-domain space.
    In the second stage, motivated by the insight that the misalignments in pixel-level can be eliminated to a certain extent in feature-level, we design an unsupervised image reconstruction network to eliminate the artifacts from features to pixels.
    Specifically, the reconstruction network can be implemented by a low-resolution deformation branch and a high-resolution refined branch, learning the deformation rules of image stitching and enhancing the resolution simultaneously.
    %To establish an evaluation benchmark and train the learning framework, a comprehensive real-world image dataset for unsupervised deep image stitching is presented and we release it at \url{www.github.com/nie-lang/UnsupervisedDeepImageStitching}.
    To establish an evaluation benchmark and train the learning framework, a comprehensive real-world image dataset for unsupervised deep image stitching is presented and released $\footnote{\url{www.github.com/nie-lang/UnsupervisedDeepImageStitching}}$.
    Extensive experiments well demonstrate the superiority of our method over other state-of-the-art solutions. Even compared with the supervised solutions, our image stitching quality is still preferred by users.

    % Traditional feature-based image stitching technologies rely heavily on feature detection quality, often failing to stitch images with few features or at low resolution.
    % The learning-based image stitching solutions are rarely studied because of the lack of real stitched labels, making the supervised methods unreliable.
    % In this paper, we propose the first unsupervised deep image stitching framework that can be achieved in two stages: unsupervised coarse image alignment and unsupervised image reconstruction. In the first stage, we design an ablation-based loss to constrain the unsupervised deep homography network, which is more suitable for large-baseline scenes than the existing constraints. Then, a transformer layer is implemented to align the input images in the stitching-domain space. In the next stage, we design an unsupervised image reconstruction network consists of a low-resolution deformation branch and a high-resolution refined branch to learn the deformation rules of image stitching and enhance the resolution of stitched results at the same time, eliminating artifacts by reconstructing the stitched images from feature to pixel.
    % Also, a comprehensive real dataset for unsupervised deep image stitching, which can be available at \url{www.github.com/nie-lang/UnsupervisedDeepImageStitching}, is proposed to evaluate our algorithms. Extensive experiments well demonstrate the superiority of our method over other state-of-the-art solutions. Even compared with the supervised solutions, our image stitching quality is still preferred by users.
\end{abstract}
\begin{IEEEkeywords}
    Computer vision, deep image stitching, deep homogrpahy estimation
\end{IEEEkeywords}

%\markboth{IEEE TRANSACTIONS ON IMAGE PROCESSING}
\markboth{}
{Shell \MakeLowercase{\textit{et al.}}: Bare Demo of IEEEtran.cls for IEEE Transactions on Magnetics Journals}
\IEEEpeerreviewmaketitle

\section{Introduction}
\label{section1}

\begin{figure}[!t]
    \centering
    {\includegraphics[width=0.48\textwidth]{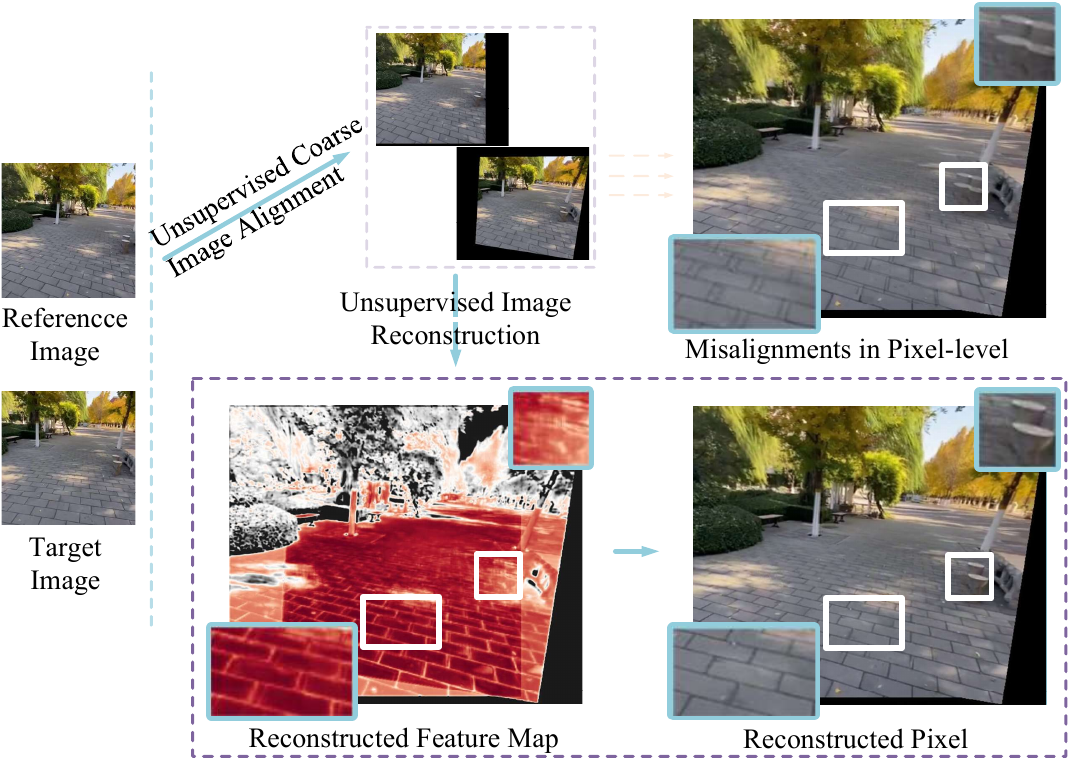}}
    \vspace{-0.4cm}
    \caption{The pipeline of proposed unsupervised deep image stitching. In the coarse alignment stage, the inputs are warped using a single homography. In the reconstruction stage, the warped images are used for reconstructing the stitched image from feature to pixel.}
    %\label{fig:long}
    %\label{fig:onerow}
    \label{Reconstruction-based}
 \end{figure}

\setlength{\epigraphwidth}{8cm}
\epigraph{Remember that all models are wrong; the practical question is how wrong do they have to be to not be useful.}%
         {George E. P. Box}

\IEEEPARstart {I}MAGE stitching is a crucial and challenging computer vision task that has been well-studied in the past decades, with the purpose to construct a panorama with a wider field-of-view (FOV) from different images captured from different viewing positions. This technology can be of great use in varying fields such as biology \cite{chalfoun2017mist, semenishchev2017method}, medical \cite{li2017medical}, surveillance videos \cite{li2019attentive, gaddam2016tiling}, autonomous driving \cite{wang2020multi, lai2019video}, virtual reality (VR) \cite{anderson2016jump, kim2019deep}.

Conventional image stitching solutions are feature-based methods, where feature detection is the first step that can profoundly affect stitching performance. Then a parametric image alignment model can be established using the matched features, by which we can warp the target image to align with the reference image. Finally, the stitched image can be obtained by assigning pixel values to each pixel in overlapping areas between the warped images.

Among these steps, establishing a parametric image alignment model is crucial in the feature-based methods. In fact, the homography transformation is the most used image alignment model, which contains translation, rotation, scaling, and vanishing point transformation, accounting for the transformation from one 2D plane to another \cite{hartley2003multiple} correctly.
However, each image domain may contain multiple different depth levels in actual scenes, which contradicts the planar scene assumption of the homography. There are often ghosting effects in the stitched results since a single homography cannot account for all the alignments at different depth levels.

Conventional feature-based solutions alleviate the artifacts in two mainstream ways. The first way is to eliminate the artifacts by aligning the target image with the reference image as much as possible \cite{gao2011constructing, lin2011smoothly, zaragoza2013projective, chang2014shape, chang2012line, chen2016natural, lin2015adaptive, li2017parallax, lee2020warping, li2019local}. These methods partition an image into different areas and compute the homography matrix for each diverse area. By exerting spatially-varying warpings on these areas, the overlapping areas are well aligned, and the artifacts are significantly reduced. The second way is to hide the artifacts by researching for an optimal seam to stitch the warped images \cite{eden2006seamless, zhang2014parallax, lin2016seagull, gao2013seam, hejazifar2018fast, 1608144}. Through optimizing a seam-related cost, the overlapping can be divided into two complementary regions along the seam. Then, a stitched image
is formed according to two regions. The feature-based solutions can significantly reduce the artifacts in most scenes. Still, they rely heavily on feature detection so that the stitching performance can drop sharply or even fail in scenes with few features or at low resolution.
\begin{figure}[!t]
    \centering
    {\includegraphics[width=0.47\textwidth]{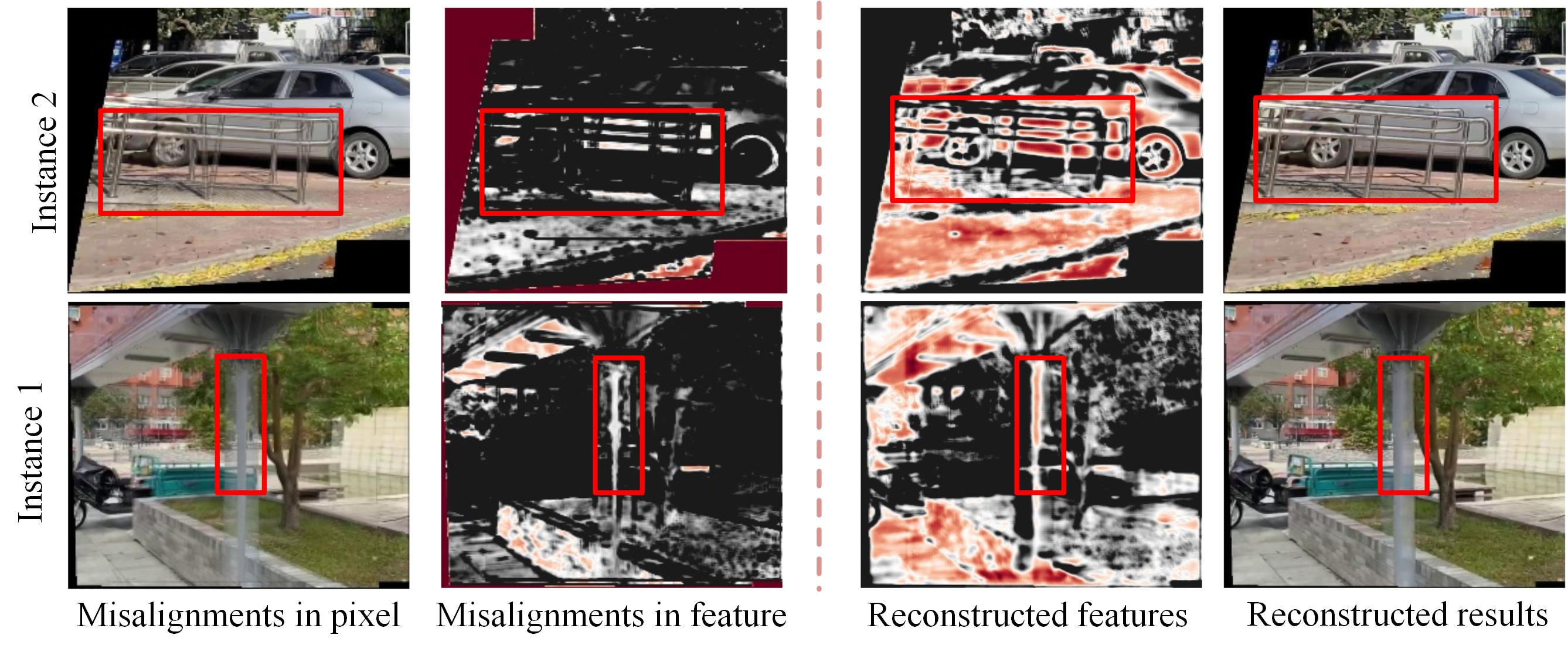}}
    \vspace{-0.4cm}
    \caption{Motivation: the misalignments in pixel-level can be visually weakened in feature-level. Col 1: the results of stitching the warped images from unsupervised coarse alignment stage. Col 2: the results of stitching the warped features extracted by the `conv1\_2' in VGG19 \cite{simonyan2014very}. Col 3-4: reconstructing from feature to pixel by unsupervised reconstruction network.}
    %\label{fig:long}
    %\label{fig:onerow}
    \label{misalignment}
 \end{figure}

Due to the incredible feature extraction capability of Convolutional Neural Networks (CNNs), recently learning-based approaches have achieved state-of-the-art performance in various fields such as depth estimation \cite{8331148}, optical flow estimation \cite{sun2018pwc, 9159910}, distortion rectification \cite{8962122}. Increasing researchers try to apply CNNs to image stitching. In \cite{hoang2020deep, shi2020image}, the CNNs are only used to extract feature points, while in \cite{lai2019video, shen2019real, li2019attentive}, the CNNs are proposed to stitch images with fixed viewing positions. Regrettably, these methods are either not a complete learning-based framework \cite{hoang2020deep, shi2020image}, or can only be used to stitch images with fixed views instead of arbitrary views \cite{lai2019video, shen2019real, li2019attentive}. Then, view-free deep image stitching methods \cite{nie2020view, nie2020learning} are proposed to overcome the two problems simultaneously. In these view-free solutions, deep image stitching can be completed by a deep homography module, a spatial transformer module, and a deep image refined module. However, all the solutions are supervised methods, and there is no real dataset for deep image stitching because of the unavailability of stitched labels in actual scenes until now. Therefore, these networks can only be trained on a `no-parallax' synthetic dataset, resulting in unsatisfying applications in real scenes.

To overcome the limitations of feature-based solutions and supervised deep solutions, we propose an unsupervised deep image stitching framework that comprises an unsupervised coarse image alignment stage and an unsupervised image reconstruction stage. The pipeline is shown in Fig.  \ref{Reconstruction-based}. In the first stage, we coarsely align the input images using a single homography.
%Most of the existing unsupervised deep homography methods require extra image contents around the input images as supervision \cite{nguyen2018unsupervised, zhang2019content}.  To relax this constraint, we propose an ablation-based loss to optimize our unsupervised deep homography network that is more suitable for the large-baseline scenes, \uline{where large-baseline is a relative concept to small-baseline in}\cite{zhang2019content}.
Different from the existing unsupervised deep homography solutions \cite{nguyen2018unsupervised, zhang2019content} that require extra image contents around the input images as supervision, we design an ablation-based loss to optimize our unsupervised deep homography network that is more suitable for the large-baseline scenes, where large-baseline is a relative concept to small-baseline in\cite{zhang2019content}.
%Besides, a stitching-domain transformer layer that is improved from Spatial Transformer Layer \cite{jaderberg2015spatial}, is proposed to warp the input images in the stitching-domain with less occupied space, compared with the existing view-free deep image stitching \cite{nie2020view, nie2020learning}.
Besides, a stitching-domain transformer layer is proposed to warp the input images in the stitching-domain with less occupied space than the existing deep stitching works \cite{nie2020view, nie2020learning}.
%In the second stage, we did not further align the warped images by optimizing a mesh, nor did we search for an optimal seam to hide artifacts. Instead, we present a novel strategy to generate the stitched images from feature to pixel, eliminating the artifacts by unsupervised image reconstruction.
In the second stage, we present an ingenious strategy to reconstruct the stitched images from feature to pixel, eliminating the artifacts by unsupervised image reconstruction.
In particular, we design a low-resolution deformation branch and a high-resolution refined branch in the reconstruction network to learn the deformation rules of image stitching and enhances the resolution, respectively.

This reconstruction strategy is motivated by an observation: misalignments in feature-level are more unnoticeable than in pixel-level (Fig. \ref{misalignment} left). Compared with pixels, feature maps are more blurred, which indicates the misalignments in pixel-level can be eliminated to a certain extent in feature-level. Therefore, we believe it is easier to eliminate artifacts in feature-level than in pixel-level. To implement this, we first reconstruct the features of the stitched image that are as close to the two warped images as possible (Col 3 in Fig. \ref{misalignment}). Then the stitched image can then be reconstructed at pixel-level (Col 4 in Fig. \ref{misalignment}) based on the reconstructed features.

The existing dataset in learning-based solutions \cite{nie2020view, nie2020learning} is a `no-parallax' synthetic dataset that cannot represent the practical application scene. And the datasets in feature-based solutions are too few to support deep learning training. To enable our framework the generalization ability in real scenarios, we also propose a large real-world image stitching dataset
containing varying overlap rates, varying degrees of parallax, and variable scenes such as indoor, outdoor, night, dark, snow, and zooming. Here, we define overlap rate as the percentage of the overlapping area in the total area of the image.

In experiments, we evaluate our performance in homography estimation and image stitching. Experimental results demonstrate the superiority of our method over other state-of-the-art solutions in real scenes.
%Even if compared with the supervised image stitching solutions, the results of our unsupervised approach are still preferred by users in terms of visual quality.
The contributions of this paper are summarized as follows:

\begin{itemize}
    \item We present an unsupervised deep image stitching framework consisting of an unsupervised coarse image alignment stage and an unsupervised image reconstruction stage.
    \item We propose the first large real dataset for unsupervised deep image stitching (to the best of our knowledge), which we hope can work as a benchmark dataset and promote other related research work.
    \item Our algorithm outperforms the state-of-the-art, including homography estimation solutions and image stitching solutions in real scenes. Even compared with the supervised solutions, our image stitching quality is still preferred by users.
 \end{itemize}

% The remainder of this paper is organized as follows: In section \ref{section2}, we review the related work of our work. Section \ref{section3} and section \ref{section4} elaborates the details of unsupervised coarse image alignment stage and unsupervised image reconstruction stage, respectively. The dataset and experiments are shown in section \ref{section5}. Finally, we conclude our work in section \ref{section7}.

\section{Related work}
\label{section2}
In this section, we subsequently review the existing works in image stitching and deep homography estimation.

\subsection{Feature-Based Image Stitching}
\label{section21}
%Feature-based image stitching methods have been well-studied in the past decades, emerging abundant excellent algorithms.
According to different strategies to eliminate artifacts, the feature-based image stitching algorithms can be divided into the following two categories:

\begin{spacing}{1.5}
\end{spacing}
\noindent\textbf{Adaptive Warping Methods.} Considering that a single transformation model is not enough to accurately align images with parallax, the idea of combining multiple parametric alignment models to align the images as much as possible is introduced. In \cite{gao2011constructing}, the dual-homography warping (DHW) is presented to align the foreground and the background, respectively. This method works well in the scene composed of two predominating planes but shows poor performance in more complex scenes. Lin $et\ al.$ \cite{lin2011smoothly} apply multiple smoothly varying affine (SVA) transformations in different regions, enhancing local deformation and alignment performance. Zaragoza $et\ al.$ \cite{zaragoza2013projective} propose the as-projective-as-possible (APAP) approach, where an image can be partitioned into dense grids, and each grid would be allocated a corresponding homography by weighting the features. In fact, APAP would still exhibit parallax artifacts in the vicinity of the object boundaries, for dramatic depth changes might occur in these areas. To get rid of this problem, the warping residual vectors are proposed to distinguish matching features from different depth planes in \cite{lee2020warping}, contributing to more naturally stitched images.

\begin{spacing}{1.5}
\end{spacing}
\noindent\textbf{Seam-Driven Methods} Seam-driven image stitching methods are also influential, acquiring natural stitched images by hiding the artifacts. Inspired by the idea of interactive digital photomontage \cite{agarwala2004interactive}, Gao $et\ al.$ \cite{gao2013seam} propose to choose the best homography with the lowest seam-related cost from candidate homography matrices. Then the artifacts are hidden through seam cutting. Referring to the optimization strategy of content-preserving warps (CPW) \cite{liu2009content}, Zhang and Liu \cite{zhang2014parallax} propose a seam-based local alignment approach while maintaining the global image structure using an optimal homography. This work was also extended to stereoscopic image stitching \cite{zhang2015casual}. Using the iterative warp and seam estimation, Lin $et\ al.$ \cite{lin2016seagull} find the optimal local area to stitch images, which can protect the curve and line structure during image stitching.

\begin{spacing}{1.5}
\end{spacing}
These feature-based algorithms contribute to perceptually nature stitched results. However, they rely heavily on the quality of feature detection, often failing in scenes with few features or at low resolution.

\subsection{Learning-Based Image Stitching}
\label{section22}
Getting a real dataset for stitching is difficult. In addition, deep stitching is quite challenging for the scenes with low overlap rate and large parallax. Subjected to these two problems, learning-based image stitching is still in development.
\begin{spacing}{1.5}
\end{spacing}
\noindent\textbf{View-Fixed Methods.}
View-fixed image stitching methods are task-driven, which are designed for the specific application scenes such as autonomous driving \cite{wang2020multi, lai2019video}, surveillance videos \cite{li2019attentive}. In these works, the end-to-end networks are proposed to stitch images from fixed views while they cannot be extended to stitch images from arbitrary views.

\begin{spacing}{1.5}
\end{spacing}
\noindent\textbf{View-Free Methods.}
To stitch images from arbitrary views using CNNs, some researchers propose to adopt CNNs in the stage of feature detection \cite{hoang2020deep, shi2020image}. However, these methods can not be regarded as a complete learning-based framework strictly. The first complete learning-based framework to stitch images from arbitrary views was proposed in \cite{nie2020view}. The images can be stitched through three stages: homography estimation, spatial transformation, and content refinement.
%Nevertheless, this work cannot stitch images with an arbitrary size,
Nevertheless, this work cannot handle input images with arbitrary resolutions due to the fully connected layers in the network,
and the stitching quality in real applications is unsatisfying. Following this deep stitching pipeline, an edge-preserved deep image stitching solution was proposed in \cite{nie2020learning}, freeing the limitation of input resolution and significantly improving the stitching performance in real scenes. %Nevertheless, this solution often causes discontinuities at the edges while eliminating artifacts.

%To train this network, a synthetic image stitching dataset is also proposed. However, this work can stitch synthetic images very well while showing disappointing performance in stitching real images.
%Then, an edge-preserved image stitching approach was proposed in \cite{nie2020learning}, contributing to perceptually natural stitched results even in real image stitching scenes.

\subsection{Deep Homography Schemes}
\label{section23}
The first deep homography method was put forward in \cite{detone2016deep}, where a VGG-style \cite{simonyan2014very} network was used to predict the eight offsets of four vertices of an image, thus uniquely determine a corresponding homography. Nguyen $et\ al.$ \cite{nguyen2018unsupervised} proposed the first unsupervised deep homography approach with the same architecture as \cite{detone2016deep} with an effective unsupervised loss. Introducing spatial attention to deep homography network, Zhang $et\ al.$ \cite{zhang2019content} proposes a content-aware unsupervised network, contributing to SOTA performance in small-baseline deep homography. In \cite{le2020deep}, multi-scale features are extracted to predict the homography from coarse to fine using image pyramids.

Besides that, the deep homography network is usually adopted as a part of the view-free image stitching frameworks \cite{nie2020view, nie2020learning}. Different from \cite{detone2016deep, nguyen2018unsupervised, zhang2019content, le2020deep}, the deep homography in image stitching is more challenging, for the baseline between input images is usually 2X$\sim$3X larger. %And it is the homography estimation capability in a large baseline that is the most crucial for obtaining a good stitched result.

%------------------------------------------------------------------------
\section{Unsupervised Coarse Image Alignment}
\label{section3}

Given two high-resolution input images, we first estimate the homography using a deep homography network in an unsupervised manner. Then the input images can be warped to align each other coarsely in the proposed stitching-domain transformer layer.

\subsection{Unsupervised Homography}
\label{section31}

\begin{figure}[!t]
    \centering
    % {\includegraphics[width=0.5\textwidth]{./figures/fig3.pdf}}
    % \vspace{-0.7cm}
    \subfigure[A failure case of padding-based strategy.]
    {\includegraphics[width=0.23\textwidth]{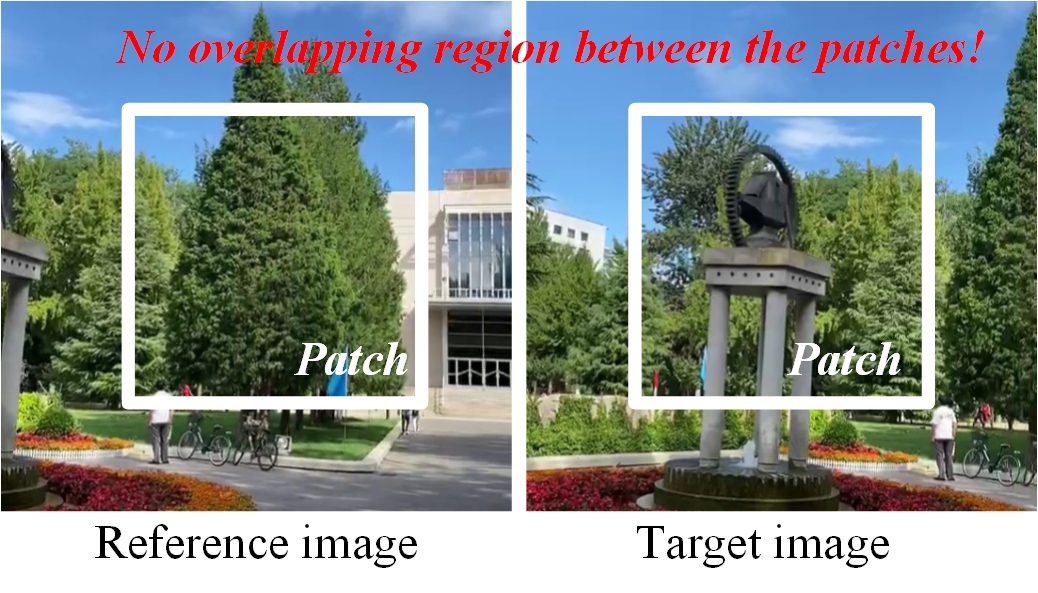}}
    \subfigure[The proposed ablation-based strategy.]
    {\includegraphics[width=0.236\textwidth]{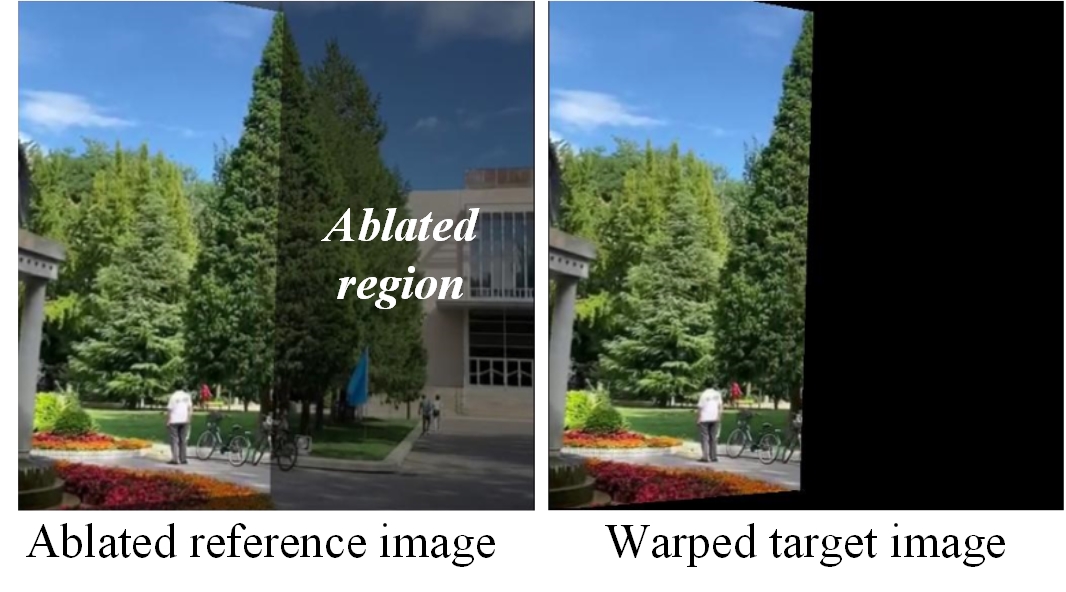}}
    \vspace{-0.2cm}
    \caption{An instance to show that the proposed ablation-based strategy is more suitable for large-baseline unsupervised homography estimation.}%Left: a failure case of padding-based strategy.. Right: the proposed ablation-based strategy.}
    %\label{fig:long}
    %\label{fig:onerow}
    \label{Unsupervised_homography}
 \end{figure}

%The unsupervised homography is the first step of our unsupervised image stitching framework, estimating a homography from the overlapping areas of two images.
The existing unsupervised deep homography methods \cite{nguyen2018unsupervised, zhang2019content} take the image patches as the input, which is shown in the white squares in Fig. \ref{Unsupervised_homography} (a). The objective function of these methods can be expressed as Eq. \eqref{EQ_padding-based}:
\begin{equation}
    L_{PW} = \left \|  \mathcal{P}(I^A)-\mathcal{P}(\mathcal{H}(I^B))   \right \|_1,
    \label{EQ_padding-based}
 \end{equation}
where $I^A, I^B$ represent the full images of the reference image and the target image, respectively. $\mathcal{P}(\cdot)$ is the operation of extracting an image patch from a full image, and $\mathcal{H}(\cdot)$ warps one image to align with the other using estimated homography. From Eq. \eqref{EQ_padding-based}, we can see that to make the warped target patch close to the reference patch, the extra contents around the target patch are utilized to pad the invalid pixels in the warped target patch. We call it a padding-based constraint strategy. This strategy works well in small-baseline \cite{zhang2019content}, or middle-baseline \cite{nguyen2018unsupervised} homography estimations while it fails in the large-baseline case. In particular, when the baseline is too large (as illustrated in Fig. \ref{Unsupervised_homography} (a)), there might be no overlapping area between the input patches, which leads to the meaningless estimation of homography from these patches.

To solve this problem, we design an ablation-based strategy to constrain large-baseline unsupervised homography estimation. Specifically, we take the full images as the input, ensuring that all overlapping areas are included in our inputs. When we enforce the warped target image close to the reference image, we no longer pad the invalid pixels in the warped image. Instead, we ablate the contents in the reference image where the invalid pixels in the warped target image locate, as shown in Fig. \ref{Unsupervised_homography} (b). Our objective function for unsupervised homography is formulated as Eq. \eqref{EQ_ablation-based}:
\begin{equation}
    L_{PW}^{'} = \left \|  \mathcal{H}(E)\odot I^A-\mathcal{H}(I^B)   \right \|_1,
    \label{EQ_ablation-based}
 \end{equation}
 where $\odot$ is the pixel-wise multiplication and $E$ is an all-one matrix with identical size with $I^A$.

 As for the architecture of our unsupervised homography network, we adopt a multi-scale deep model proposed in \cite{nie2020learning},
 which connects feature pyramid and feature correlation in a unified framework so that it can predict the honography from coarse to fine and handle relative large-baseline scenes.

 %as it has achieved excellent performance in the large-baseline supervised homography estimation.

\subsection{Stitching-Domain Transformer Layer}
\label{section32}

\begin{figure}[!t]
    \centering
    \subfigure[]
    {\includegraphics[width=0.35\textwidth]{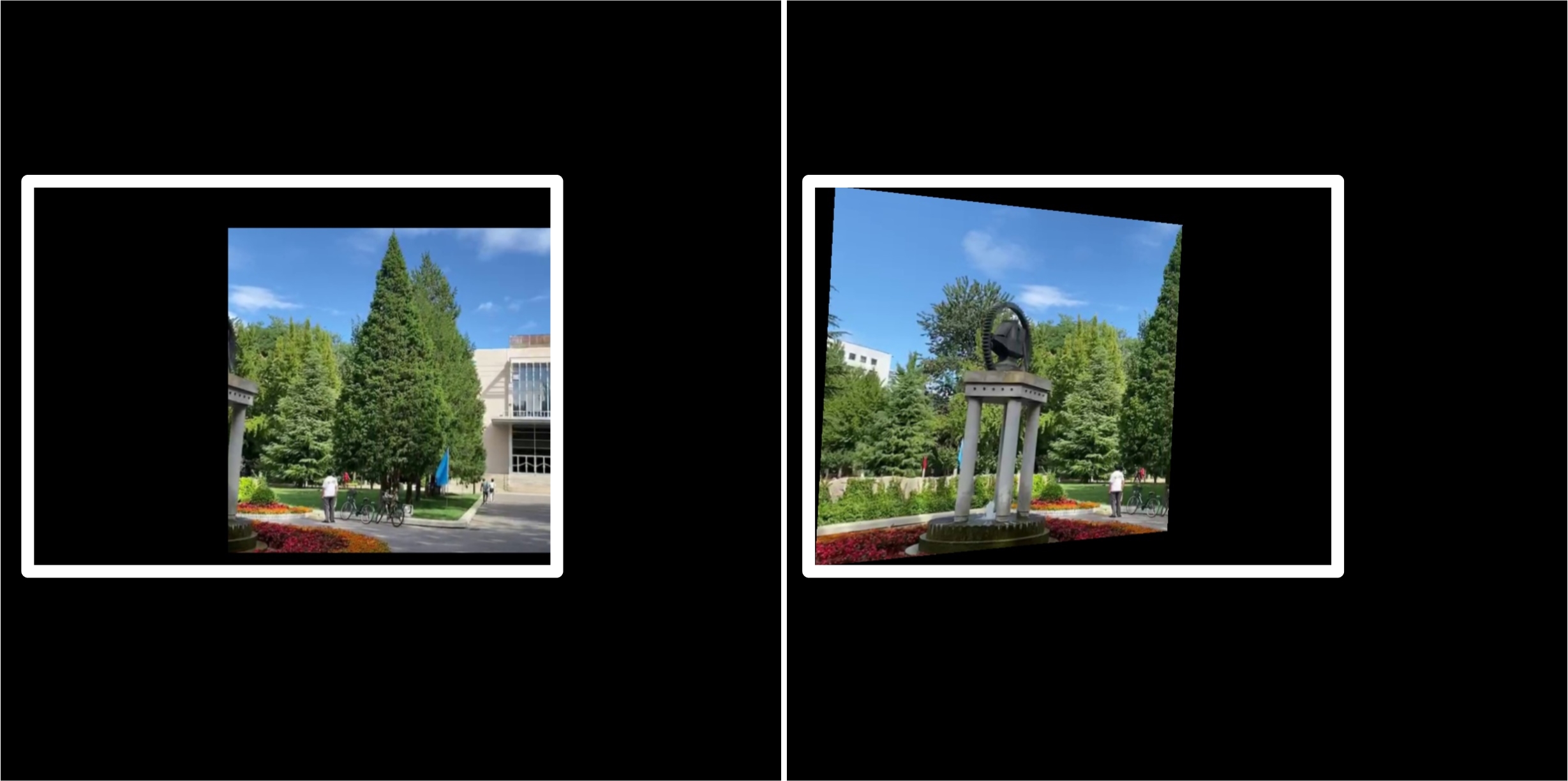}}
    \subfigure[]
    {\includegraphics[width=0.12\textwidth]{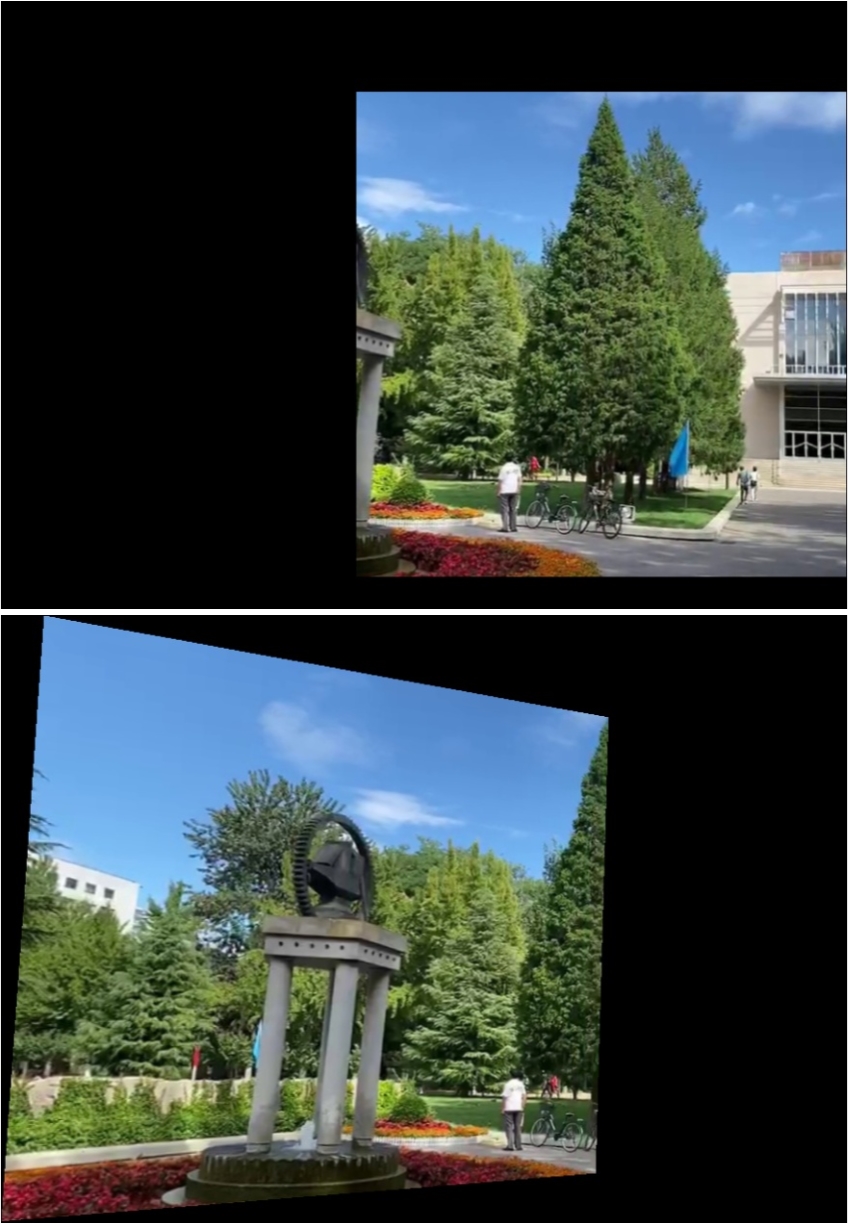}}
    \vspace{-0.2cm}
    \caption{The comparison between the spatial transformer layer in existing deep image stitching and our stitching-domain transformer layer. (a): Warping by spatial transformer layer in existing deep image stitching \cite{nie2020view, nie2020learning}. (b): Warping by our stitching-domain transformer layer.}
    %\label{fig:long}
    %\label{fig:onerow}
    \label{Transformer_layer}
 \end{figure}

 The spatial transformer layer was first proposed in \cite{jaderberg2015spatial}, where images can be spatially transformed with gradient backpropagation guaranteed using the homography model. In image stitching, input images of the same resolution can output stitched images of different resolution according to the varying overlapping rates, which brings a considerable challenge to deep image stitching. The existing deep image stitching methods solve this problem by extending the spatial transformer layer \cite{nie2020view, nie2020learning}. Specifically, these solutions define a maximum resolution for the stitched image so that all the input contents can be included in the output. In addition, the network will output images with the same resolution every time.
 %By defining a maximum size for the stitched results, most possible results can be completely included in.
 However, most of the space occupied by black pixels outside the white box in Fig. \ref{Transformer_layer} (a) are wasted. To deal with spatial waste, we propose a stitching-domain transformer layer. We define the stitching-domain as the smallest bounding rectangle of the stitched image, which saves the most space while ensuring the integrity of the image contents. The warped results of ours are illustrated in Fig. \ref{Transformer_layer} (b), and our stitching-domain transformer layer can be implemented as follows.

 \begin{figure*}[!t]
    \centering
    \includegraphics[width=1\textwidth]{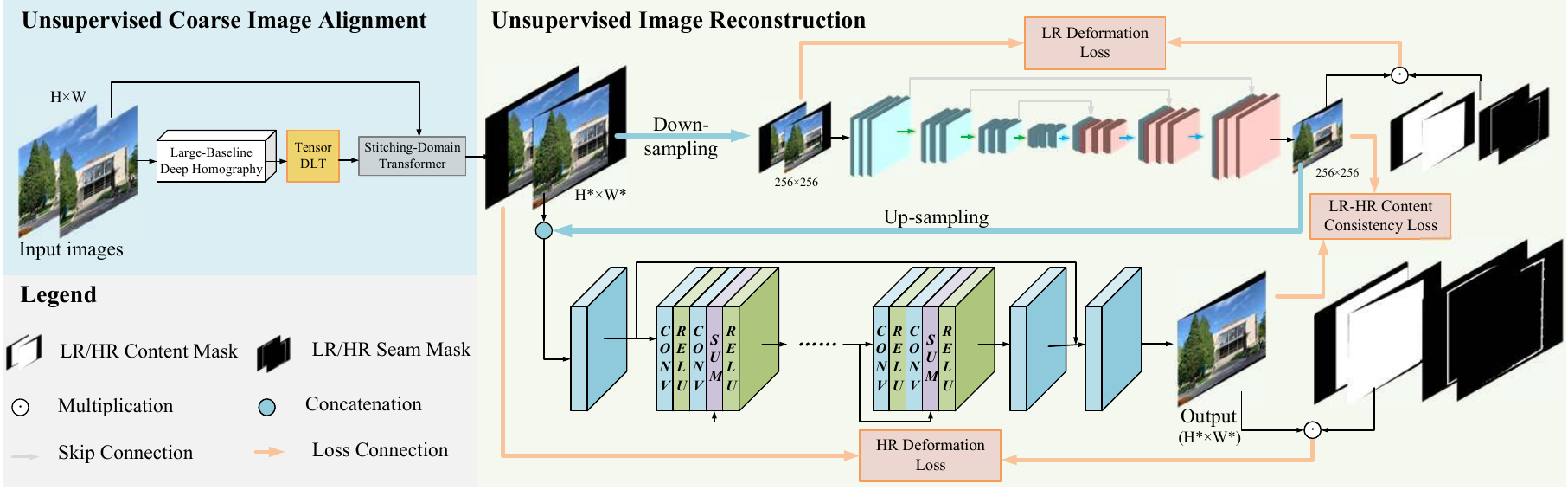}
    \vspace{-0.6cm}
    \caption{An overview of our unsupervised deep image stitching. Left: the unsupervised coarse image alignment stage. Right: the unsupervised image reconstruction stage.}
    \label{EP-Stitching}
 \end{figure*}

 First, we calculate the coordinates of the 4 vertices in the warped target image by Eq. \eqref{eq3}:
 \begin{equation}
    (x^W_k, y^W_k) =  (x^B_k, y^B_k) +  (\Delta x_k, \Delta y_k), k\in \{1,2,3,4\},
    \label{eq3}
 \end{equation}
where $(x^W_k, y^W_k)$, $(x^B_k, y^B_k)$ are the $k$-th vertex coordinates of the warped target image and the target image, respectively. $(\Delta x_k, \Delta y_k)$ donate the offsets of the $k$-th vertex that are estimated form the aforementioned homogrpahy network. Then, the size of the warped image ($H^*\times W^*$) can be obtained by Eq. \eqref{eq4}:

 \begin{equation}
    \begin{aligned}
    W^*&= \max \limits_{k\in \{1,2,3,4\}}\{x^W_k, x^A_k \} - \min \limits_{k\in \{1,2,3,4\}}\{x^W_k, x^A_k \}, \\
    H^*&= \max \limits_{k\in \{1,2,3,4\}}\{y^W_k, y^A_k \} - \min \limits_{k\in \{1,2,3,4\}}\{y^W_k, y^A_k \},
    \end{aligned}
    \label{eq4}
 \end{equation}
where $(x^A_k, y^A_k)$ are the vertex coordinates of the reference image that have the same values as $(x^B_k, y^B_k)$.
Finally, we assign the specific values to the pixels of the warped images ($I^{AW}, I^{BW}$) from the input images ($I^A, I^B$), which can be represented as Eq. \eqref{eq5}:

\begin{equation}
    \begin{aligned}
    I^{AW}&=  \mathcal{W}(I^A, I),\\
    I^{BW}&=  \mathcal{W}(I^B, H),
    \end{aligned}
    \label{eq5}
 \end{equation}
where $I$ and $H$ are the identity matrix and the estimated homography matrix, respectively. And $\mathcal{W}(\cdot)$ donates the operation of warping an image using a $3\times 3$ transformation matrix with the stitching-domain set to $H^*\times W^*$.

In this way, we transform the input images in the stitching-domain space, effectively reducing the space occupied by feature maps in the subsequent reconstruction network. Compared with the transformer layer used in \cite{nie2020learning,nie2020view}, the proposed layer can help to stitch larger resolution images when the GPU memory is limited.

\section{Unsupervised Image Reconstruction}
\label{section4}
Considering the limitation that a single homography can only represent the spatial transformation in the same depth \cite{hartley2003multiple}, the input images cannot be completely aligned in the real-world dataset in the first stage.
To break the bottleneck of single homography, we propose to reconstruct the stitched image from feature to pixel.
The overview of the proposed unsupervised deep image stitching framework is illustrated in Fig. \ref{EP-Stitching}. The reconstruction network can be implemented by two branches: low-resolution deformation branch (Fig. \ref{EP-Stitching} top) and high-resolution refined branch (Fig. \ref{EP-Stitching} bottom), learning the deformation rules of image stitching and enhancing the resolution, respectively.

%To avoid insufficient receptive field caused by high resolution, the deformation rules of image stitching are first learned in low-resolution. Moreover, the high-resolution branch is designed to enhance the resolution of the stitched image.

\subsection{Low-Resolution Deformation Branch}
\label{section41}

\begin{figure}[!t]
    \centering
    {\includegraphics[width=0.5\textwidth]{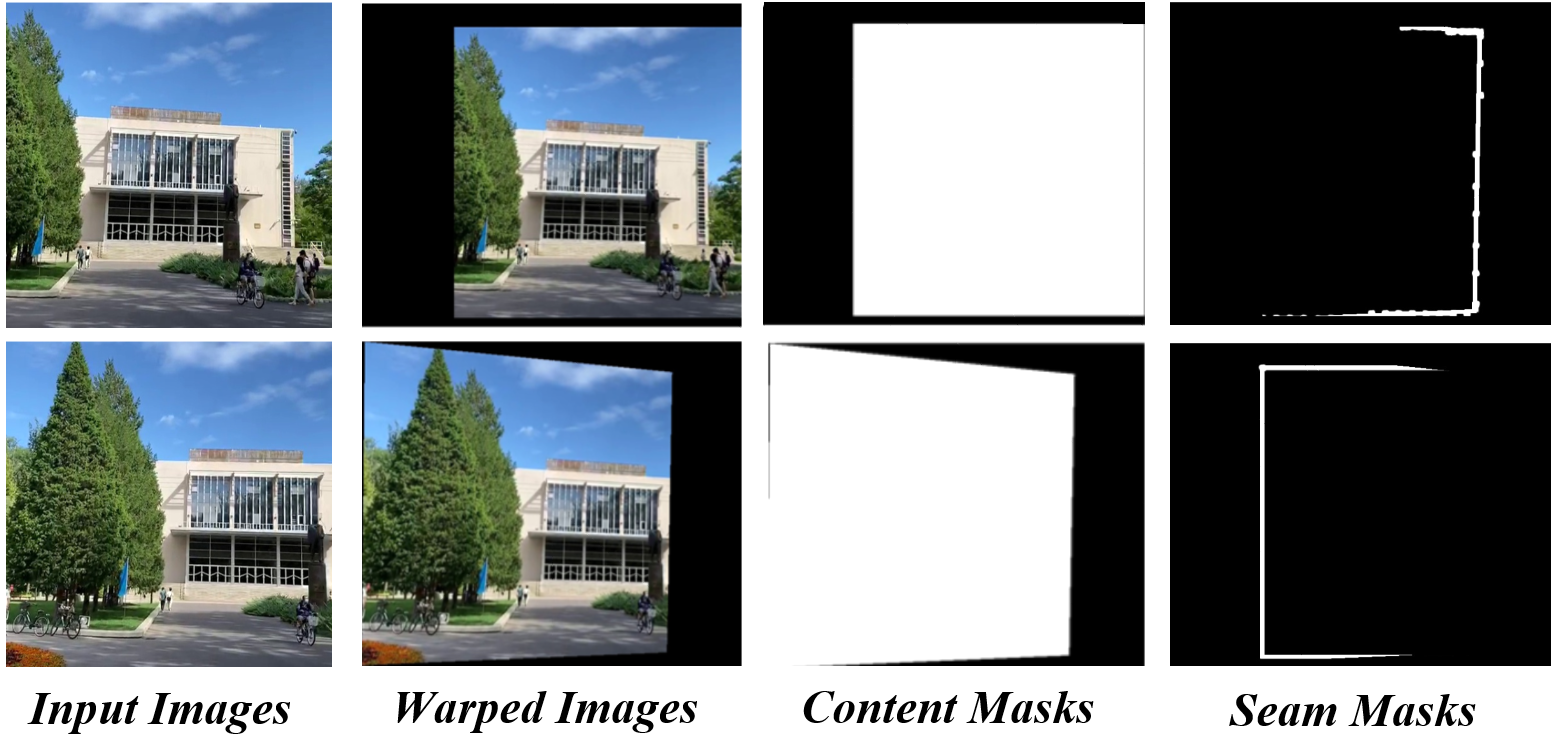}}
    \vspace{-0.6cm}
    \caption{Learning deformation rules with masks in low-resolution. From left to right, each column represents input images ($I^{A}, I^{B}$), low-resolution warped images ($I^{AW}, I^{BW}$), content masks ($M^{AC}, M^{BC}$), and seam masks ($M^{AS}, M^{BS}$).}
    %\label{fig:long}
    %\label{fig:onerow}
    \label{masks}
 \end{figure}

%In fact, since the input images contain different degrees of parallax and the estimated homography cannot be perfectly accurate inevitably, the warped images obtained in the first stage cannot be completely aligned. The second stage is designed to eliminate artifacts by reconstructing the stitched images from feature to pixel.
Reconstructing the images only in the high-resolution branch is not appropriate because the receptive field decreases relatively as the resolution increases.
To ensure that the receptive field of the network can completely perceive misaligned regions (especially in the case of high resolution and large parallax), we designed a low-resolution branch to learn the deformation rules of image stitching first.
%To ensure a sufficiently large field of reception during the learning process, we learn the deformation rules of image stitching in low-resolution first.
%As shown in Fig. \ref{EP-Stitching} (top), the warped images are first down-sampled to $256\times 256$.
As shown in Fig. \ref{EP-Stitching}(top), the warped images are first down-sampled to a low-resolution, defined as 256$\times$256, in our implementation.
Then an encoder-decoder network consisting of 3 pooling layers and 3 deconvolutional layers is used to reconstruct the stitched image. The filter numbers of the convolutional layers are set to 64, 64, 128, 128, 256, 256, 512, 512, 256, 256, 128, 128, 64, 64, and 3, respectively. Furthermore, skip connections are adopted to connect the low-level and high-level features with the same resolution \cite{ronneberger2015u}.

\begin{figure}[!t]
    \centering
    {\includegraphics[width=0.45\textwidth]{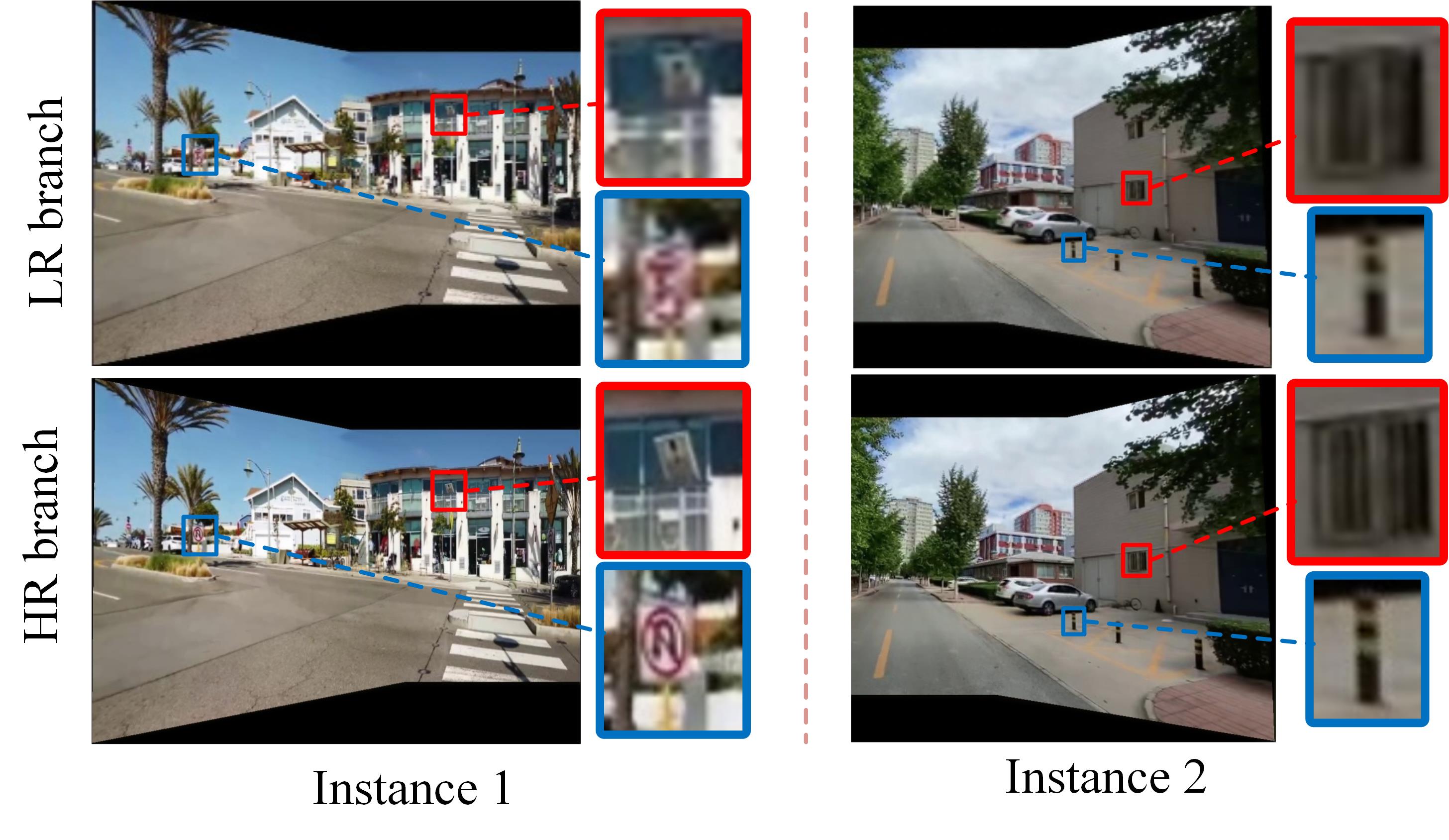}}
    \vspace{-0.4cm}
    \caption{The outputs of the low-resolution branch and high-resolution branch. The high-resolution branch is designed to enhance the resolution and refine the stitched image.}
    %\label{fig:long}
    %\label{fig:onerow}
    \label{HR_function}
 \end{figure}

 \begin{figure*}[!t]
    \centering
    {\includegraphics[width=1\textwidth]{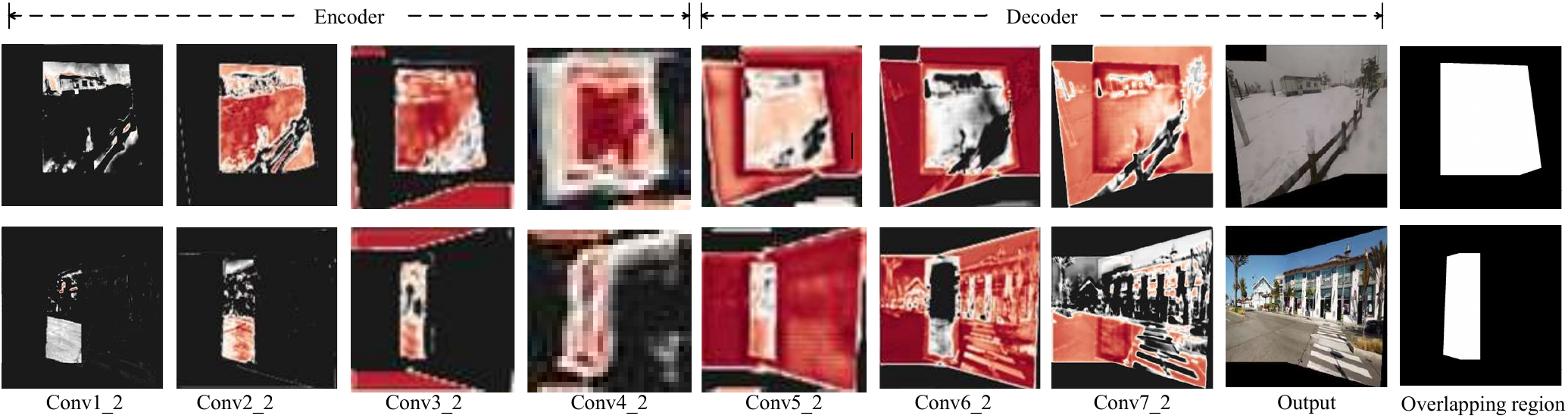}}
    \vspace{-0.8cm}
    \caption{Visualization of the learning process of the low-resolution deformation branch. The stitched images are reconstructed from overlapping regions to non-overlapping regions.}
    %\label{fig:long}
    %\label{fig:onerow}
    \label{feature_map}
 \end{figure*}

In this process, the deformation rules of image stitching are learned with content masks and seam masks (Fig. \ref{masks}). The content masks are adopted to constrain the features of the reconstructed image close to the warped images, while the seam masks are designed to constrain the edges of the overlapping areas to be natural and continuous. In particular, we obtain the content masks ($M^{AC}, M^{BC}$) using Eq. \eqref{eq5} by replacing the $I^{A}, I^{B}$ with an all-one matrix $E_{H\times W}$, and the seam masks can be calculated by Eq. \eqref{nabla} and Eq. \eqref{eq_seam}:
\begin{flalign}
    %\begin{aligned}
    \nabla M^{AC}&=  |M^{AC}_{i,j}-M^{AC}_{i-1,j}|+|M^{AC}_{i,j}-M^{AC}_{i,j-1}|,\nonumber\\
    \nabla M^{BC}&=  |M^{BC}_{i,j}-M^{BC}_{i-1,j}|+|M^{BC}_{i,j}-M^{BC}_{i,j-1}|,\label{nabla}\\
    M^{AS}&=  \mathcal{C}(\nabla M^{BC}*E_{3\times 3}*E_{3\times 3}*E_{3\times 3}) \odot M^{AC},\nonumber\\
    M^{BS}&=  \mathcal{C}(\nabla M^{AC}*E_{3\times 3}*E_{3\times 3}*E_{3\times 3}) \odot M^{BC},\label{eq_seam}
    %\end{aligned}
 \end{flalign}
where $(i,j)$ donates the coordinate location, $*$ represents the operation of convolution, and $\mathcal{C}$ clips all the elements to between 0 and 1.
%, and $E_{3\times 3}$ is a $3\times 3$ convolutional filter with all values set to 1.
Then we design the content loss and seam loss in low-resolution as Eq. \eqref{loss_lr_content} and Eq. \eqref{loss_lr_seam}:

\begin{equation}
    \begin{aligned}
    \mathcal{L}^{l}_{Content} = &\mathcal{L}_{P}(S_{LR}\odot M^{AC}, I^{AW})\\
    +&\mathcal{L}_{P}(S_{LR}\odot M^{BC}, I^{BW}), \\
    \end{aligned}
    \label{loss_lr_content}
\end{equation}

\begin{equation}
    \begin{aligned}
    \mathcal{L}^{l}_{Seam} = &\mathcal{L}_{1}(S_{LR}\odot M^{AS}, I^{AW}\odot M^{AS})\\
    +&\mathcal{L}_{1}(S_{LR}\odot M^{BS}, I^{BW}\odot M^{BS})\\
    \end{aligned}
    \label{loss_lr_seam}
\end{equation}
where $S_{LR}$ is the low-resolution stitched image. $\mathcal{L}_{1}$ and $\mathcal{L}_{P}$ donate the L1 loss and the perceptual loss \cite{johnson2016perceptual}, respectively. To make the feature of the reconstructed image as close to that of the warped images as possible, we calculate the perceptual loss on layer `conv5\_3' of VGG-19 \cite{simonyan2014very} which is deep enough to shrink the feature difference between the warped images. Next, the total loss function of low-resolution unsupervised deformation can be formulated as Eq. \eqref{loss_lr}:

\begin{equation}
    \begin{aligned}
        \mathcal{L}_{LR} = \lambda_c\mathcal{L}^{l}_{Content} + \lambda_s\mathcal{L}^{l}_{Seam}
    \end{aligned}
    \label{loss_lr}
\end{equation}
where $\lambda_s$ and $\lambda_c$ weight the contribution of the content constraint and seam constraint.

\subsection{High-Resolution Refined Branch}
\label{section42}
After the initialized deformation in the low-resolution branch, we develop a high-resolution refined branch to enhance the resolution and refine the stitched image.
The high-resolution refers to the resolution of the output of the first stage. Actually, in our dataset, the resolution is bigger than 512$\times$512.
To illustrate the effect of high-resolution branch, we exhibit the outputs of two branches in Fig. \ref{HR_function}.
This branch is composed of convolutional layers entirely, as shown in Fig. \ref{EP-Stitching} (bottom), which means it can deal with pictures of arbitrary resolution. To be specific, it consists of three separate convolutional layers and eight resblocks \cite{he2016deep}, of which the filter number of each layer is set to 64 except that of the last layer is set to 3. To prevent low-level information from being gradually forgotten as the convolutional network gets deep, the feature of the first layer is added with that of the penultimate layer. Moreover, each resblock is composed of convolution, relu, convolution, sum, and relu.

We up-sample $S_{LR}$ to the resolution of the warped images and concatenate them together as the input of this branch. The output is the high-resolution stitched image $S_{HR}$. And we conclude the loss function of the high-resolution refined branch $\mathcal{L}_{HR}$ imitating Eq. \eqref{loss_lr} as Eq. \eqref{loss_hr}:
\begin{equation}
    \begin{aligned}
        \mathcal{L}_{HR} = \lambda_c\mathcal{L}^{h}_{Content} + \lambda_s\mathcal{L}^{h}_{Seam}
    \end{aligned}
    \label{loss_hr}
\end{equation}
where $\mathcal{L}^{h}_{Content}$ and $\mathcal{L}^{h}_{Seam}$ are the content loss and seam loss in high-resolution which can be calculated using Eq. \eqref{loss_lr_content}, \eqref{loss_lr_seam} by replacing the $S_{LR}$ and low-resolution masks with the $S_{HR}$ and the high-resolution masks.
When calculating the $\mathcal{L}_{P}$ in high resolution, we adopt the layer `conv3\_3' of VGG-19, since this layer is shallower than the layer `conv5\_3' (used in $\mathcal{L}_{P}$ of low resolution) and the output using this layer is more clear.

\subsection{Objective Function}
\label{section43}
The high-resolution branch is designed to refine the stitched image, but it tends to cause artifacts in the stitched image, since the increase in resolution can relatively reduce the receptive field of the network (more details can be found in Section \ref{section54}). To enable our network the abilities to enhance resolution and to eliminate parallax artifacts simultaneously, a content consistency loss is proposed as Eq. \eqref{content_consistency}:
\begin{equation}
    \mathcal{L}_{CS} = \left \| S_{HR}^{256\times 256}-S_{LR}\right \|_1,
    \label{content_consistency}
\end{equation}
where $S_{HR}^{256\times 256}$ is obtained by resizing $S_{HR}$ to $256\times 256$ that is the resolution of the output in low-resolution branch.

Taking all the constraints into consideration, we conclude our objective function of the image reconstruction stage as Eq. \eqref{objective_function}:
\begin{equation}
    \mathcal{L}_{R} = \omega_{LR}\mathcal{L}_{LR} +  \omega_{HR}\mathcal{L}_{HR} +  \omega_{CS}\mathcal{L}_{CS},
    \label{objective_function}
\end{equation}
where the $\omega_{LR}$, $\omega_{HR}$ and $\omega_{CS}$ represent weights of each part.

\subsection{Reconstruction from Feature to Pixel}
\label{section44}
To exhibit the learning process from feature to pixel, we visualized the feature maps of the low-resolution deformation branch in Fig. \ref{feature_map}. At the very beginning of the encoder stage, the network only focuses on the overlapping areas, and the features of non-overlapping areas are all suppressed. Next, as the resolution decreases, deeper semantic features are extracted and reconstructed. In the decoder stage, the network begins to pay attention to non-overlapping areas besides overlapping areas. As the resolution is restored, clearer feature maps are reconstructed. Finally, the stitched image is reconstructed at the pixel level.

%------------------------------------------------------------------------
\section{Experiments}
\label{section5}
% \begin{figure}[!t]
%     \centering
%     {\includegraphics[width=0.5\textwidth]{./figures/dataset.jpg}}
%     \vspace{-0.8cm}
%     \caption{Illustrations of our proposed unsupervised deep image stitching dataset.}
%     %\label{fig:long}
%     %\label{fig:onerow}
%     \label{dataset}
%  \end{figure}

 \begin{figure}[!t]
    \centering
    \subfigure[Varying scenes in our dataset.]
    {\includegraphics[width=0.47\textwidth]{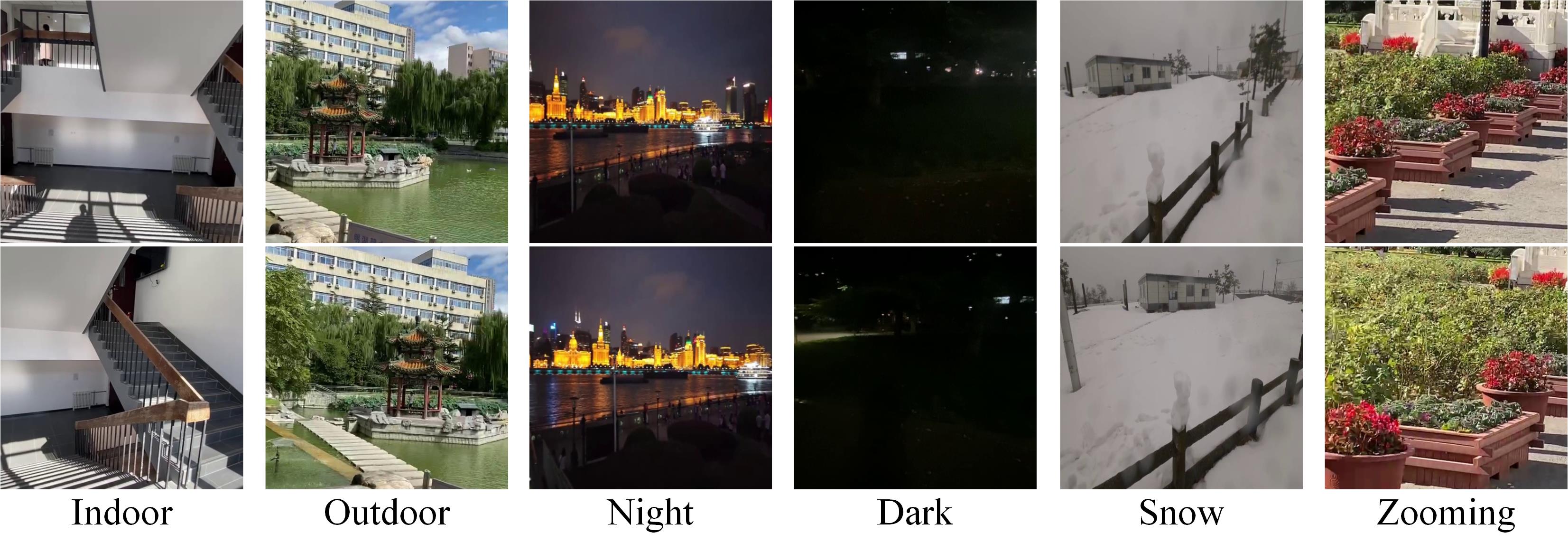}}
    \subfigure[Varying overlap rates in our dataset.]
    {\includegraphics[width=0.47\textwidth]{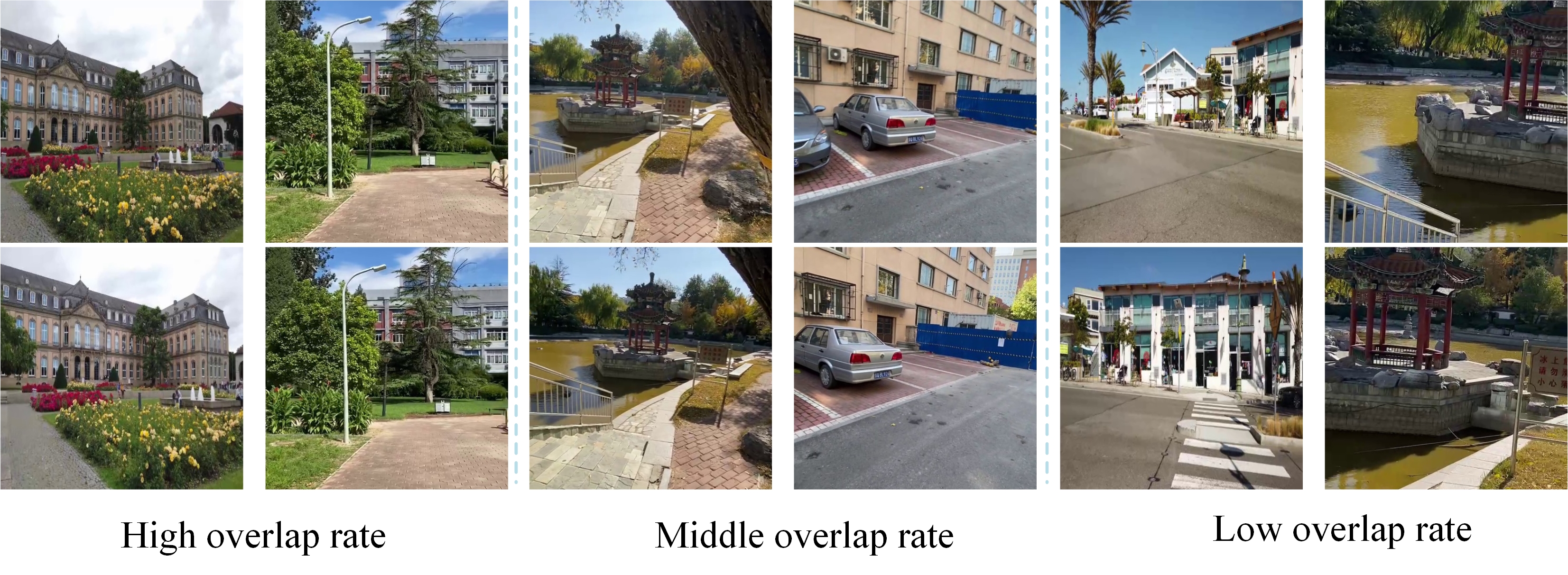}}
    \subfigure[Varying degrees of parallax in our dataset.]
    {\includegraphics[width=0.47\textwidth]{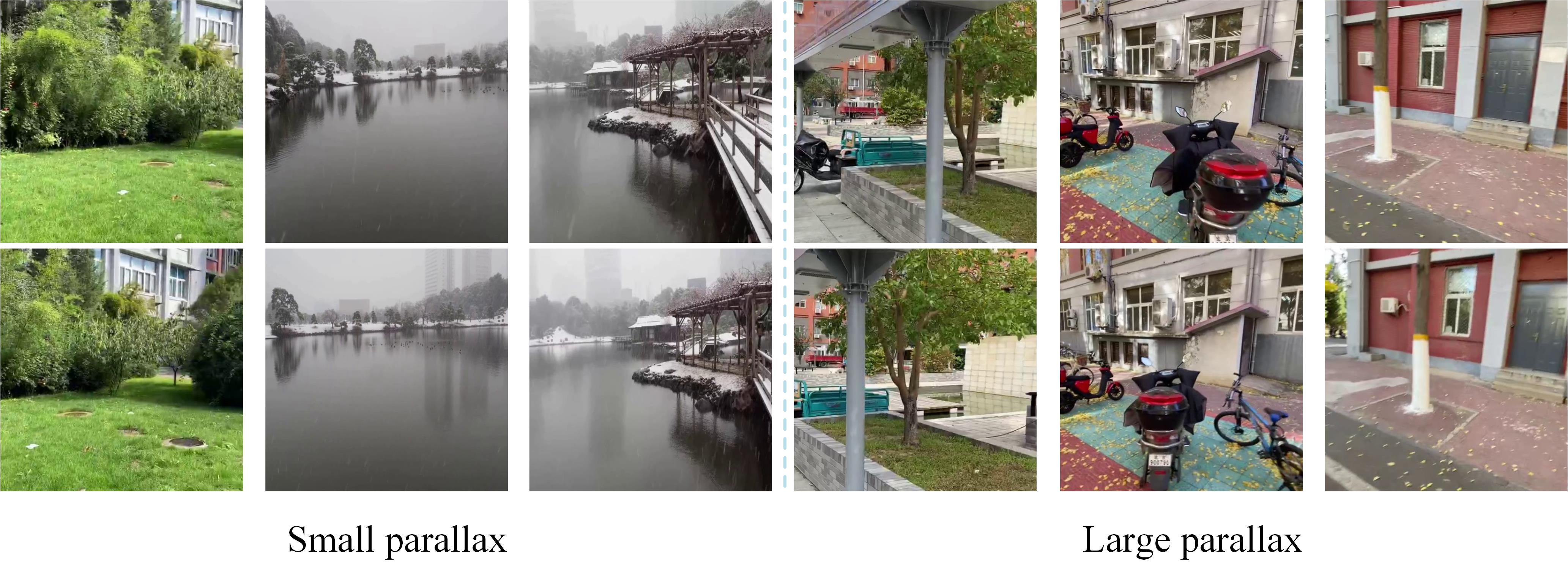}}
    \vspace{-0.2cm}
    \caption{Illustrations of our proposed unsupervised deep image stitching dataset.}
    %\label{fig:long}
    %\label{fig:onerow}
    \label{dataset}
 \end{figure}

 In this section, extensive experiments are conducted to validate effectiveness of the proposed method.

\subsection{Dataset and Implement Details}
\label{section51}
\noindent\textbf{Dataset.}
To train our network, we also propose an unsupervised deep image stitching dataset that is obtained from variable moving videos. Of these videos, some are from \cite{zhang2019content} and the others are captured by ourselves. By extracting the frames from these videos with different interval time, we get the samples with different overlap rates (Fig. \ref{dataset} (b)). Moreover, these videos are not captured by the camera rotating around the optical center, and the shot scenes are far from a planar structure, which means this dataset contains different degrees of parallax (Fig. \ref{dataset} (c)). Besides, this real-world dataset includes variable scenes such as indoor, outdoor, night, dark, snow, and zooming (Fig. \ref{dataset} (a)).

To quantitatively describe the distribution of different overlap rates and varying degrees of parallax in our dataset. We divide the overlap rates into 3 levels and define a high overlap rate greater than 90\%, a middle overlap rate ranging from 60\%-90\%, and a low overlap rate lower than 60\%. This classification criterion is formulated according to \cite{zhang2019content,detone2016deep,nguyen2018unsupervised}, where \cite{zhang2019content} is the represnetative work in high overlap rate. The average overlap rate of the proposed dataset is greater than 90\%. And \cite{detone2016deep, nguyen2018unsupervised} are the representative works in middle overlap rate for the average overlap rate of Warped COCO (disturbance $<$ 32) dataset \cite{detone2016deep} is about 75\%. Besides, to describe parallax accurately, we align the target image with the reference image using a global homography and then calculate the maximum misalignment error of corresponding feature points in the coarse aligned images to show the magnitude of parallax. In this way, we divide the parallax into 2 levels: small parallax with error smaller than 30 pixels and large parallax with error greater than 30 pixels. Fig. \ref{dataset} (c) demonstrates the difference of different parallax intuitively.

In particular, we get 10,440 cases for training and 1,106 for testing. Among our dataset, the ratios of overlap rates from high to low are about 16\%, 66\%, and 18\%, while the ratios of parallax from small to large are about 91\% and 9\%.
Although our dataset contains no ground-truth, we include our testing results in this dataset, which we hope can work as a benchmark dataset for other researchers to follow and compare.

\begin{spacing}{1.5}
\end{spacing}
\noindent\textbf{Details.}
We train our unsupervised image stitching framework in three steps. First, we train our deep homography network on the synthetic dataset (Stitched MS-COCO\cite{nie2020view}) for 150 epochs. Second, we finetune the homography network on the proposed real dataset for 50 epochs. Third, we train the deep image reconstruction network on the proposed real dataset for 20 epochs. All the training process is unsupervised, which means our framework only takes the reference/target image as input and requires no label. The optimizer is Adam \cite{kingma2014adam} with an exponentially decaying learning rate with an initial value of $10^{-4}$. We set $\lambda_s$ and $\lambda_c$ to 2 and $10^{-6}$. And $\omega_{LR}$, $\omega_{HR}$ and $\omega_{CS}$ are set to 100, 1 and 1, respectively. In testing, it takes about 0.4$s$ to stitch 2 input images with resolution of 512$\times$512. All the components of this framework are implemented on TensorFlow. Both the training and testing are conducted on a single GPU with NVIDIA RTX 2080 Ti.

\subsection{Comparison of Homography Estimation}
\label{section52}
To evaluate the performance of the proposed ablation-based unsupervised deep homography objectively, we compare our solution with $I_{3\times 3}$, SIFT\cite{lowe2004distinctive}+RANSAC\cite{fischler1981random}, DHN\cite{detone2016deep}, UDHN\cite{nguyen2018unsupervised}, CA-UDHN\cite{zhang2019content}, and LB-DHN \cite{nie2020learning} on the synthetic dataset and real dataset respectively. The $I_{3\times 3}$ refers to a $3\times 3$ identity matrix as a `no-warping’ homography for reference, and SIFT+RANSAC is chosen as the representative of traditional homography solutions because it outperforms most traditional solutions as shown in \cite{nguyen2018unsupervised, zhang2019content}. The DHN, UDHN, CA-UDHN, and LB-DHN are the deep learning solutions, of which UDHN and CA-UDHN are the unsupervised solutions that both adopt the padding-based strategy to train their networks.

\begin{table*}[]
    \centering
    \caption{Comparison experiment on homography estimation. The 1st and 2nd best solutions are marked in red and blue, respectively.}
   \subtable[4pt-Homography RMSE ($\downarrow$) on Warped MS-COCO (synthetic)]{
    \scalebox{0.8}{
        \begin{tabular}{|l||l|l||l|l||l|l|l|}
            \hline
            \makecell[c]{\multirow{2}{*}{Method}}  & \multicolumn{2}{c||}{Traditional homography} & \multicolumn{2}{c||}{Deep homography (supervised)} & \multicolumn{3}{c|}{Deep homography (unsupervised)}\\
            \cline{2-8}
            & \makecell[c]{$I_{3\times 3}$} & SIFT\cite{lowe2004distinctive}+RANSAC\cite{fischler1981random} & DHN\cite{detone2016deep} & \makecell[c]{LB-DHN \cite{nie2020learning}} & UDHN\cite{nguyen2018unsupervised}  & CA-UDHN\cite{zhang2019content}  & Ours\_v1 (synthetic) \\
            \hline \hline
            \makecell[c]{Top 0$\sim$30\%}& \makecell[c]{15.0154} & \makecell[c]{$\mathbf{\textcolor{blue}{0.6743}}$} & \makecell[c]{3.2998} & \makecell[c]{$\mathbf{\textcolor{red}{0.2719}}$} & \makecell[c]{2.1894} & \makecell[c]{15.0082} & \makecell[c]{1.1773} \\
            \hline
            \makecell[c]{30$\sim$60\%} & \makecell[c]{18.2515} & \makecell[c]{$\mathbf{\textcolor{blue}{1.0964}}$} & \makecell[c]{4.8839} & \makecell[c]{$\mathbf{\textcolor{red}{0.4140}}$} & \makecell[c]{3.5272} & \makecell[c]{18.2498}  & \makecell[c]{1.4544} \\
            \hline
            \makecell[c]{60$\sim$100\%} & \makecell[c]{21.3517} & \makecell[c]{19.0286} & \makecell[c]{7.6944} & \makecell[c]{$\mathbf{\textcolor{red}{0.9632}}$} & \makecell[c]{6.4984} & \makecell[c]{21.3618}  & \makecell[c]{$\mathbf{\textcolor{blue}{3.0702}}$} \\
            \hline \hline
            \makecell[c]{Average} & \makecell[c]{18.5220} & \makecell[c]{9.4782} & \makecell[c]{5.5358} & \makecell[c]{$\mathbf{\textcolor{red}{0.5962}}$} & \makecell[c]{4.3179} & \makecell[c]{18.5234} & \makecell[c]{$\mathbf{\textcolor{blue}{2.0239}}$} \\
      \hline
      \end{tabular}
}
\label{firsttable}
}

\qquad

\subtable[PSNR ($\uparrow$) of the overlapping regions on the proposed dataset (real)]{
    \scalebox{0.8}{
        \begin{tabular}{|l||l|l||l|l||l|l|l|}
            \hline
            \makecell[c]{\multirow{2}{*}{Method}}  & \multicolumn{2}{c||}{Traditional homography} & \multicolumn{2}{c||}{Deep homography (supervised)} & \multicolumn{3}{c|}{Deep homography (unsupervised)}\\
            \cline{2-8}
            & \makecell[c]{$I_{3\times 3}$} & SIFT\cite{lowe2004distinctive}+RANSAC\cite{fischler1981random} & DHN\cite{detone2016deep} & \makecell[c]{LB-DHN \cite{nie2020learning}} & UDHN\cite{nguyen2018unsupervised}  & Ours\_v1 (synthetic)  & Ours\_v2 (real) \\
            \hline \hline
            \makecell[c]{Top 0$\sim$30\%}& \makecell[c]{16.1923} & \makecell[c]{25.2300} & \makecell[c]{16.3957} & \makecell[c]{24.7515} & \makecell[c]{19.3851} & \makecell[c]{$\mathbf{\textcolor{blue}{26.1958}}$} & \makecell[c]{$\mathbf{\textcolor{red}{27.8386}}$} \\
            \hline
            \makecell[c]{30$\sim$60\%} & \makecell[c]{13.0546} & \makecell[c]{22.2308} & \makecell[c]{13.3648} & \makecell[c]{21.1436} & \makecell[c]{15.9251} & \makecell[c]{$\mathbf{\textcolor{blue}{22.6115}}$}  & \makecell[c]{$\mathbf{\textcolor{red}{23.9451}}$} \\
            \hline
            \makecell[c]{60$\sim$100\%} & \makecell[c]{10.8747} & \makecell[c]{17.5791} & \makecell[c]{11.5001} & \makecell[c]{18.4594} & \makecell[c]{13.1016} & \makecell[c]{$\mathbf{\textcolor{blue}{19.5277}}$}  & \makecell[c]{$\mathbf{\textcolor{red}{20.7013}}$} \\
            \hline \hline
            \makecell[c]{Average} & \makecell[c]{13.1151} & \makecell[c]{21.2541} & \makecell[c]{13.5191} & \makecell[c]{21.1418} & \makecell[c]{15.8252} & \makecell[c]{$\mathbf{\textcolor{blue}{22.4421}}$} & \makecell[c]{$\mathbf{\textcolor{red}{23.8045}}$} \\
      \hline
      \end{tabular}
}
\label{secondtable}
}

\qquad

\subtable[SSIM ($\uparrow$) of the overlapping regions on the proposed dataset (real)]{
    \scalebox{0.8}{
        \begin{tabular}{|l||l|l||l|l||l|l|l|}
            \hline
            \makecell[c]{\multirow{2}{*}{Method}}  & \multicolumn{2}{c||}{Traditional homography} & \multicolumn{2}{c||}{Deep homography (supervised)} & \multicolumn{3}{c|}{Deep homography (unsupervised)}\\
            \cline{2-8}
            & \makecell[c]{$I_{3\times 3}$} & SIFT\cite{lowe2004distinctive}+RANSAC\cite{fischler1981random} & DHN\cite{detone2016deep} & \makecell[c]{LB-DHN \cite{nie2020learning}} & UDHN\cite{nguyen2018unsupervised}  & Ours\_v1 (synthetic)  & Ours\_v2 (real) \\
            \hline \hline
            \makecell[c]{Top 0$\sim$30\%}& \makecell[c]{0.3869} & \makecell[c]{0.8598} & \makecell[c]{0.4088} & \makecell[c]{0.8249} & \makecell[c]{0.5732} & \makecell[c]{$\mathbf{\textcolor{blue}{0.8671}}$} & \makecell[c]{$\mathbf{\textcolor{red}{0.9023}}$} \\
            \hline
            \makecell[c]{30$\sim$60\%} & \makecell[c]{0.1730} & \makecell[c]{0.7662} & \makecell[c]{0.1699} & \makecell[c]{0.7124} & \makecell[c]{0.3344} & \makecell[c]{$\mathbf{\textcolor{blue}{0.7844}}$}  & \makecell[c]{$\mathbf{\textcolor{red}{0.8298}}$} \\
            \hline
            \makecell[c]{60$\sim$100\%} & \makecell[c]{0.0732} & \makecell[c]{0.5583} & \makecell[c]{0.0772} & \makecell[c]{0.5497} & \makecell[c]{0.1651} & \makecell[c]{$\mathbf{\textcolor{blue}{0.6270}}$}  & \makecell[c]{$\mathbf{\textcolor{red}{0.6846}}$} \\
            \hline \hline
            \makecell[c]{Average} & \makecell[c]{0.1969} & \makecell[c]{0.7105} & \makecell[c]{0.2042} & \makecell[c]{0.6805} & \makecell[c]{0.3379} & \makecell[c]{$\mathbf{\textcolor{blue}{0.7456}}$} & \makecell[c]{$\mathbf{\textcolor{red}{0.7929}}$} \\
      \hline
      \end{tabular}
}
\label{thirdtable}
}
\end{table*}

\begin{spacing}{1.5}
\end{spacing}
\noindent\textbf{Synthetic dataset.} The first comparative experiment is conducted on Warped MS-COCO that is the most known synthetic dataset for deep homography estimation. All the learning methods are trained on Warped MS-COCO. The results are illustrated in Table \ref{firsttable}, where `Ours\_v1' is our model trained with this dataset in an unsupervised manner. From Table \ref{firsttable}, we can observe:

(1) Ours\_v1 outperforms the existing unsupervised deep homography methods (UDHN, CA-UDHN), of which CA-UDHN is the SOTA solution in small-baseline deep homography. However, the performance of CA-UDHN in this dataset degenerates to be close to that of $I_{3\times 3}$ due to its limited receptive field.

(2) After adopting our ablation-based unsupervised loss to LB-DHN, 4pt-Homography RMSE increases, which means this loss is not suitable for this `no-parallax' synthetic dataset.

\begin{spacing}{1.5}
\end{spacing}
\noindent\textbf{Real Dataset.} Then, we carry on a comparison on the proposed real dataset, which consists of varying degrees of parallax. Since this dataset lacks ground truth, we adopt the PSNR and SSIM of the overlapping regions to evaluate the performance, which can be calculated as Eq. \eqref{EQ_psnr_ssim}:
\begin{equation}
    \begin{aligned}
    PSNR_{overlap} = \mathcal{PSNR}(\mathcal{H}(E)\odot I^A, \mathcal{H}(I^B)), \\
    SSIM_{overlap} = \mathcal{SSIM}(\mathcal{H}(E)\odot I^A, \mathcal{H}(I^B)),
    \end{aligned}
    \label{EQ_psnr_ssim}
\end{equation}
where $\mathcal{PSNR}(\cdot)$ and $\mathcal{SSIM}(\cdot)$ donates the operations of computing PSNR and SSIM between two images, respectively. We test DHN and UDHN using the public pretrained models. LB-DHN and Ours\_v1 are trained on Stitched MS-COCO \cite{nie2020view} which is similar to Warped MS-COCO with lower overlap rate. Ours\_v2 is the model of finetuning Ours\_v1 about 50 epochs on the proposed real dataset. By analyzing the results shown in Table \ref{secondtable} \ref{thirdtable}, we can conclude:

(1) The proposed unsupervised solution (Ours\_v2) outperforms all the methods, including the supervised ones in the real dataset.

(2) Although Ours\_v1 and LB-DHN are both trained on the synthetic dataset, Ours\_v1 achieves better performance under the real dataset, which indicates the proposed unsupervised loss can equip the network with better generalization ability.

\subsection{Comparison of Image Stitching}
\label{section53}
To verify our method's superiority in image stitching, we compare our method with feature-based solutions and compare with recent learning-based solutions (even if it is not fair to compare our unsupervised algorithms with the supervised ones).

\begin{spacing}{2}
\end{spacing}
\noindent\textbf{1) Compared with Feature-Based Solutions}

In this section, we choose global Homography\cite{hartley2003multiple}, APAP\cite{zaragoza2013projective}, robust ELA\cite{li2017parallax} as the representatives of feature-based solutions to compare with our algorithms. Of these methods, we implement Homography with global projective transformation, and we get the stitched results of APAP and robust ELA (adaptive warping methods) by running their open-source codes with our testing instances.
After alignment, image fusion is adopted to produce the stitched image and reduce artifacts. Specifically, we fuse the warped images with the pixel-weighted principle, assigning a relatively large weight to the pixel with a high intensity value.

\begin{spacing}{1.5}
\end{spacing}
\begin{figure}[!t]
    \centering
    {\includegraphics[width=0.45\textwidth]{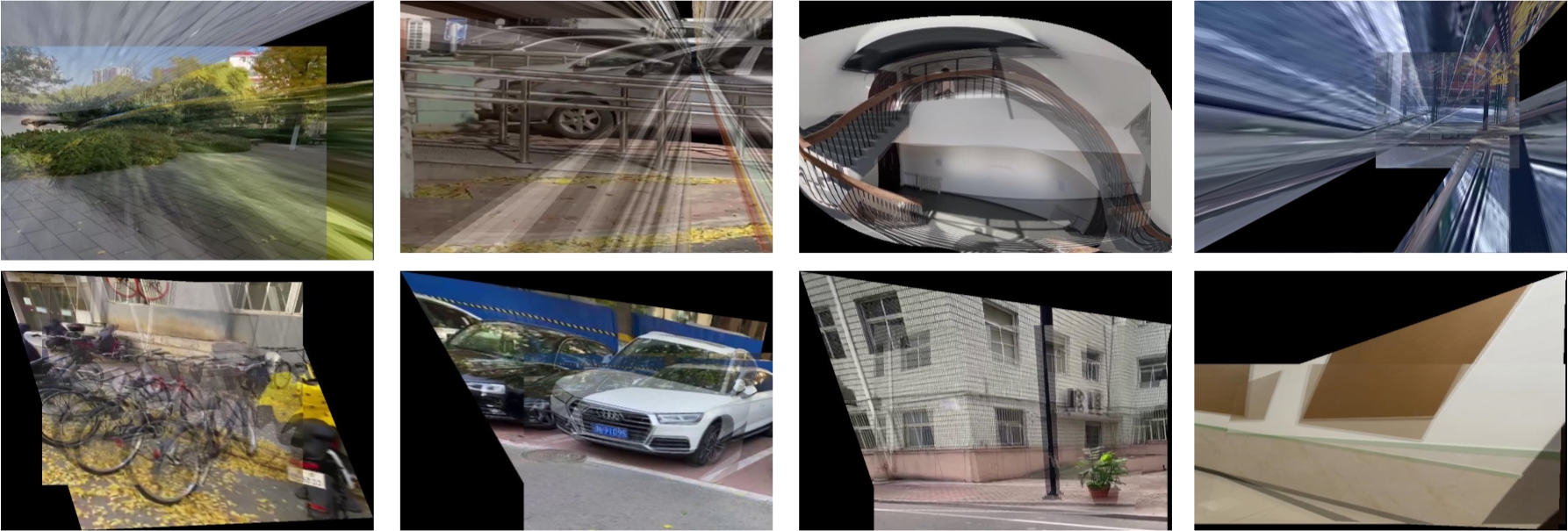}}
    \vspace{-0.2cm}
    \caption{Demonstration of `failure'. Top: significant distortion. Bottom: intolerable artifacts.}
    %\label{fig:long}
    %\label{fig:onerow}
    \label{failure}
 \end{figure}

 \begin{table*}[]
    \centering
   \caption{comparison of robustness for image stitching. The number of testing cases is 1,106.}
   \scalebox{0.75}{
    \begin{tabular}{|l|l||l|l|l||l|l||l|}
        \hline
        \multirow{2}{*}{Input resolution}  & \multirow{2}{*}{Metrics}  & \multicolumn{3}{c||}{Feature-based} & \multicolumn{2}{c||}{Learning-based (supervised)} & Learning-based (unsupervised)\\
        \cline{3-8}
              &       & Homography\cite{hartley2003multiple}  & APAP\cite{zaragoza2013projective}  & robust ELA\cite{li2017parallax} & VFISNet\cite{nie2020view} & EPISNet\cite{nie2020learning} & \makecell[c]{Ours} \\ \hline
              \makecell[c]{\multirow{4}{*}{512$\times$512}}  & Error        & \makecell[c]{0}       & \makecell[c]{3}       & \makecell[c]{0}          & \makecell[c]{-}               & \makecell[c]{0}           & \makecell[c]{0}    \\ \cline{2-8}
                                    & Failure      & \makecell[c]{86}      & \makecell[c]{31}      & \makecell[c]{111}        & \makecell[c]{-}             & \makecell[c]{22}          &  \makecell[c]{15}    \\ \cline{2-8}
                                    & Total        & \makecell[c]{86}      & \makecell[c]{34}      & \makecell[c]{111}        & \makecell[c]{-}             & \makecell[c]{22}          &    \makecell[c]{15}  \\ \cline{2-8}
                                    & Success rate & \makecell[c]{92.22\%} & \makecell[c]{96.93\%} & \makecell[c]{89.96\%}    & \makecell[c]{-}         & \makecell[c]{$\mathbf{\textcolor{blue}{98.01\%}}$}     &   \makecell[c]{$\mathbf{\textcolor{red}{98.64\%}}$}   \\ \hline \hline
                                    \makecell[c]{\multirow{4}{*}{256$\times$256}}  & Error        & \makecell[c]{0}       & \makecell[c]{10}      & \makecell[c]{0}          & \makecell[c]{-}               & \makecell[c]{0}           & \makecell[c]{0}    \\ \cline{2-8}
                                    & Failure      & \makecell[c]{88}      & \makecell[c]{40}      & \makecell[c]{124}        & \makecell[c]{-}             & \makecell[c]{22}          &   \makecell[c]{15}   \\ \cline{2-8}
                                    & Total        & \makecell[c]{88}      & \makecell[c]{50}      & \makecell[c]{124}        & \makecell[c]{-}             & \makecell[c]{22}          &   \makecell[c]{15}   \\ \cline{2-8}
                                    & Success rate & \makecell[c]{92.04\%} & \makecell[c]{95.48\%} & \makecell[c]{88.79\%}    & \makecell[c]{-}         & \makecell[c]{$\mathbf{\textcolor{blue}{98.01\%}}$}     &    \makecell[c]{$\mathbf{\textcolor{red}{98.64\%}}$}  \\ \hline \hline
                                    \makecell[c]{\multirow{4}{*}{128$\times$128}} & Error        & \makecell[c]{1}       & \makecell[c]{158}     & \makecell[c]{9}          & \makecell[c]{0}               & \makecell[c]{0}           & \makecell[c]{0}    \\ \cline{2-8}
                                    & Failure      & \makecell[c]{206}     & \makecell[c]{66}      & \makecell[c]{214}        & \makecell[c]{131}             & \makecell[c]{32}          &   \makecell[c]{15}   \\ \cline{2-8}
                                    & Total        & \makecell[c]{207}     & \makecell[c]{224}     & \makecell[c]{223}        & \makecell[c]{131}             & \makecell[c]{32}          &   \makecell[c]{15}   \\ \cline{2-8}
                                    & Success rate & \makecell[c]{81.28\%} & \makecell[c]{79.75\%} & \makecell[c]{79.84\%}    & \makecell[c]{88.16\%}         & \makecell[c]{$\mathbf{\textcolor{blue}{97.11\%}}$}      &   \makecell[c]{$\mathbf{\textcolor{red}{98.64\%}}$}   \\ \hline
    \end{tabular}
   }
    \label{robustness}
\end{table*}

% \begin{figure}[!t]
%     \centering
%     {\includegraphics[width=0.45\textwidth]{./figures/robust_fig.jpg}}
%     \vspace{-0.4cm}
%     \caption{An extremely challenging sample to compare the robustness more intuitively. The resolution of the inputs is $512\times 512$. The left column displays the original images. Since they are too dark, we impose image augmentation to better exhibit these results (right column).}
%     %\label{fig:long}
%     %\label{fig:onerow}
%     \label{robust_fig}
%  \end{figure}

 \begin{figure}[!t]
    \centering
    {\includegraphics[width=0.45\textwidth]{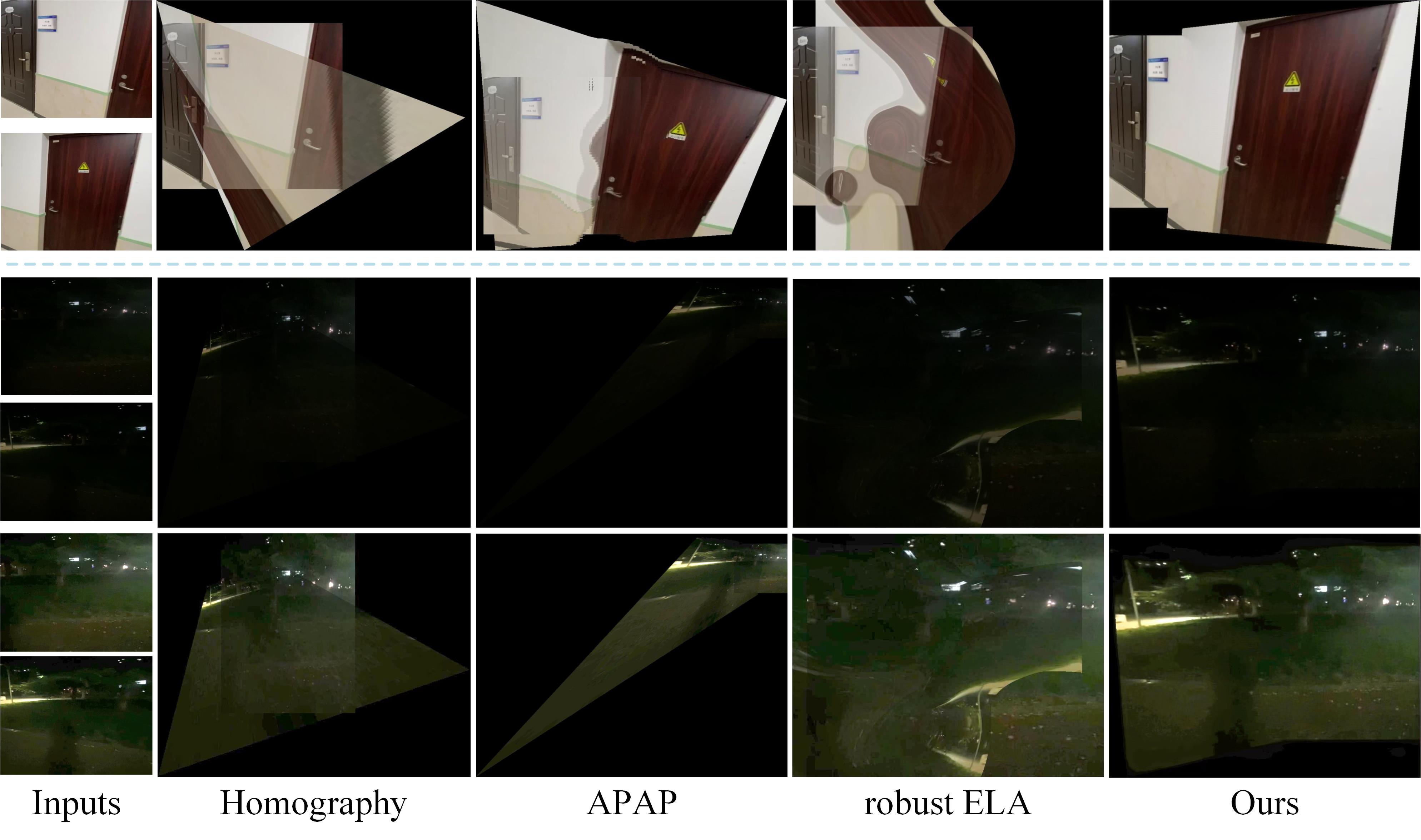}}
    \vspace{-0.4cm}
    \caption{Challenging samples to compare the robustness more intuitively in the scenes of indoors and dark. Row 1: indoors. Row 2: dark. Row 3: image augmentation to the dark scene. The resolution of the inputs is $512\times 512$.}
    %\label{fig:long}
    %\label{fig:onerow}
    \label{robust_fig}
 \end{figure}
\textbf{Study on Robustness.}
The performance of feature-based solutions is easily affected by the quantity and distribution of the feature points, resulting in weak robustness in varying scenes. By contrast, the proposed method overcomes this problem. To validate this view, we test the feature-based methods and ours on our test set (1,106 samples). To simulation the change of feature quantity, we resize the test set to different resolutions, e.g., $512\times 512$, $256\times 256$, and $128\times 128$. As the resolution decreases, the number of features decreases exponentially. The results are shown in Table \ref{robustness}, where `error' indicates the number of program crashes and `failure' refers to the number of stitching unsuccessfully. Specifically, we define significant distortion (Fig. \ref{failure} top) and intolerable artifacts (Fig. \ref{failure} bottom) as `failure'. All the stitched results of these methods will be public with our dataset. Comparing the success rates in Table \ref{robustness}, we can observe:

(1) Ours is more robust than the feature-based methods. In fact, the `error' and `failure' cases of the feature-based solutions are mainly distributed in low-light and indoor scenes, while ours performed well in these challenging scenes.

(2) As the resolution decreases, the success rates of learning-based methods decrease while ours remains robust.

Besides, to perceive the robustness more intuitively, Fig. \ref{robust_fig} demonstrates two challenging examples in the scenes of indoors and dark. Since the sample in dark is too dark to see clearly, we impose image augmentation to better exhibit these results (Row 3 in Fig. \ref{robust_fig}). These examples are challenging for the feature-based solutions because the features in these scenes are hard to detect. In contrast, our solution stitches them successfully due to the fantastic feature extraction capabilities of CNNs.

\begin{spacing}{1.5}
\end{spacing}

\begin{figure}[!t]
    \centering
    {\includegraphics[width=0.45\textwidth]{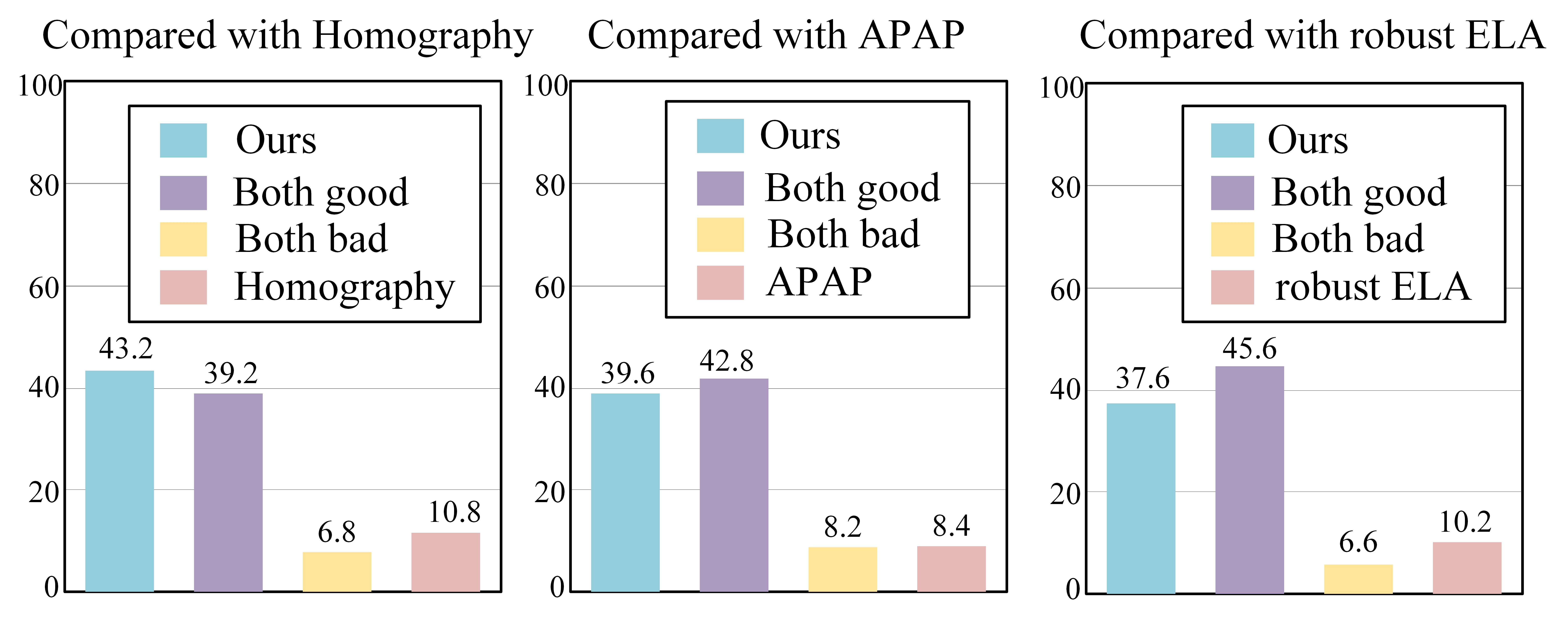}}
    \vspace{-0.4cm}
    \caption{User study on visual quality: compared with feature-based methods. The numbers are shown in percentage and averaged on 20 participants.}
    %\label{fig:long}
    %\label{fig:onerow}
    \label{user_1}
 \end{figure}

 \begin{figure*}[!t]
    \centering
    {\includegraphics[width=1\textwidth]{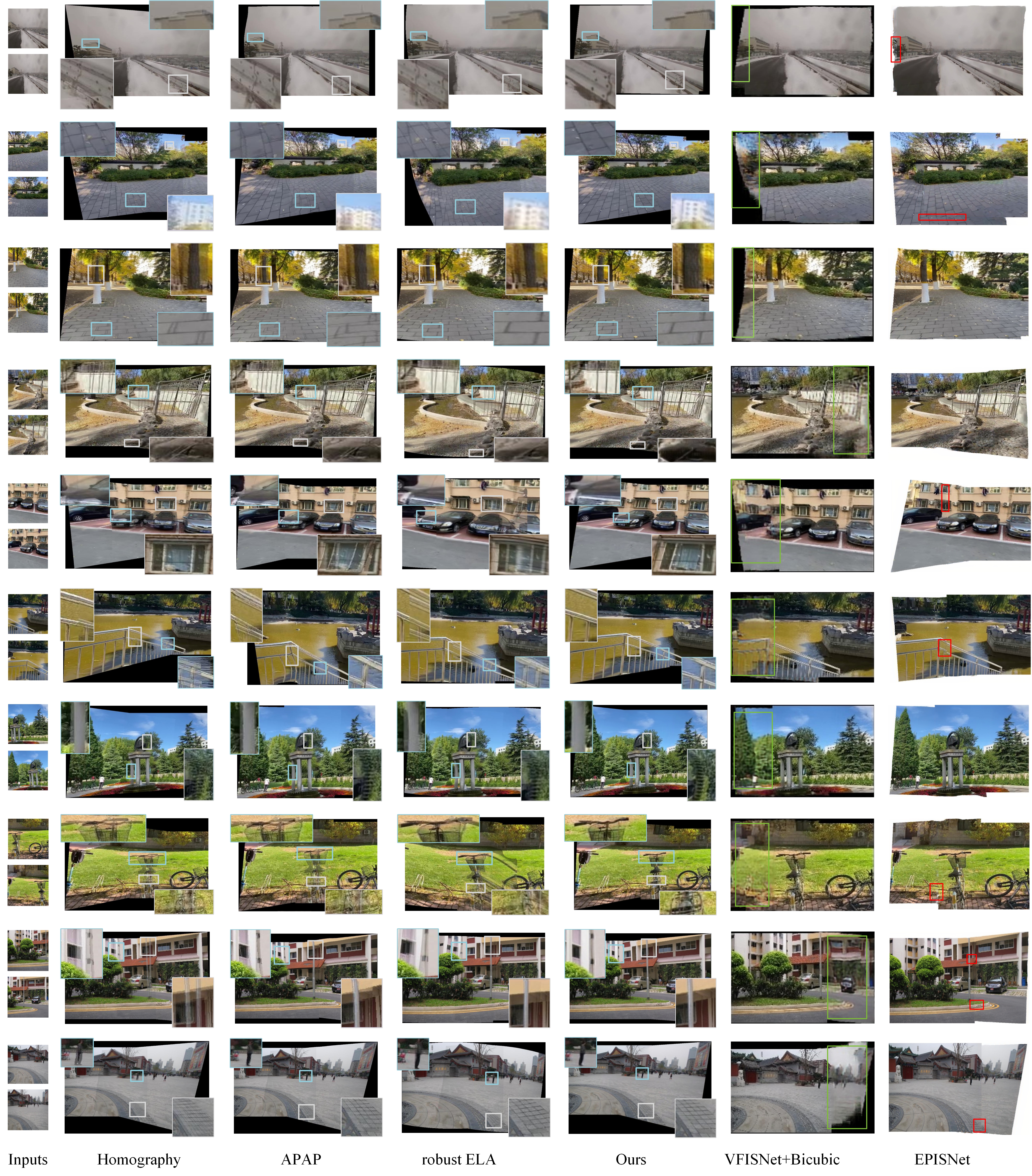}}
    \vspace{-0.8cm}
    \caption{Visual comparison of the image stitching quality. Row 1-8: instances with varying degrees of parallax from the proposed dataset. Row 9-10: ``yard"\cite{gao2013seam} and ``temple"\cite{gao2011constructing} (classic image stitching instances outside of our dataset).}

    %Col 1-2: ``yard"\cite{gao2013seam} and ``temple"\cite{gao2011constructing} (classic image stitching instances outside of our dataset). Col 3-4: ``bicycle" and ``sculpture" (instances from the proposed dataset).
    %\label{fig:long}
    %\label{fig:onerow}
    \label{real_results}
 \end{figure*}

\textbf{Study on Visual Quality.}
The proposed deep image stitching framework should be regarded as a whole which takes two images from arbitrary views as inputs and outputs the stitched result. Therefore, the traditional indicator that calculates the similarity of the overlapping regions is not suitable for our method. To compare with other methods quantitatively, we design user studies on visual quality. Specifically, we compare our method with Homography, APAP, and robust ELA one by one. At each time, four images are shown on one screen: the inputs, our stitched result, and the result from Homography/APAP/robust ELA. The results of ours and the other method are illustrated in random order each time. The user may zoom-in on the images and is required to answer which result is preferred. In the case of ``no preference," the user needs to answer whether the two results are ``both good" or ``both bad". The studies are carried out in our testing set, which means every user has to compare each method with ours in 1,106 images. In this study, we invite 20 participants, including 10 researchers/students with computer vision backgrounds and 10 volunteers outside this community.

The results are shown in Fig. \ref{user_1}. Neglecting the ratio of both good and both bad, we find that preferring ours is significantly more than preferring other methods, which means our results have higher visual quality in users' evaluation.

\begin{spacing}{1.5}
\end{spacing}
To further demonstrate our performance, we also display the stitched results on the proposed real dataset (row 1-8 in Fig. \ref{real_results}) and on the classic image stitching instances outside of our dataset (row 9-10 in Fig. \ref{real_results}). All the cases are with varying degrees of parallax.
Besides promising visual quality, it verifies the generalization ability of our model.

\begin{spacing}{2}
\end{spacing}
\noindent\textbf{2) Compared with Learning-Based Solutions}

The existing learning-based image stitching methods (VFISNet\cite{nie2020view} and EPISNet\cite{nie2020learning}) are supervised learning methods, which require extra labels to train the network. In the case that it is unfair to compare our unsupervised solution with the supervised ones, our method still exhibits a superiority over them on robustness, continuity, illumination, and visual quality.

\begin{spacing}{1.5}
\end{spacing}
\textbf{Study on Robustness.}
VFISNet is the first deep image stitching work that can stitch images from arbitrary views in a complete deep learning framework. However, it has a nonnegligible shortcoming: it can only stitch images of $128\times 128$. Therefore, only the result under the resolution of $128\times 128$ is given when measuring its robustness. The detailed results in Table \ref{robustness} shows that the robustness of ours is better than other supervised ones. This can be accounted for by the following two reasons:
(1) Our unsupervised deep homography model outperforms the other methods on robustness, which significantly reduces failure cases caused by inaccurate homography estimation.

(2) Our unsupervised deep image reconstruction model can effectively reduce artifacts by reconstructing the stitched image from feature to pixel, which reduces failure cases caused by intolerant artifacts.

\begin{figure}[!t]
    \centering
    \subfigure[Comparison of edge continuity. Left: EPISNet\cite{nie2020learning}. Right: ours.]
    {\includegraphics[width=0.45\textwidth]{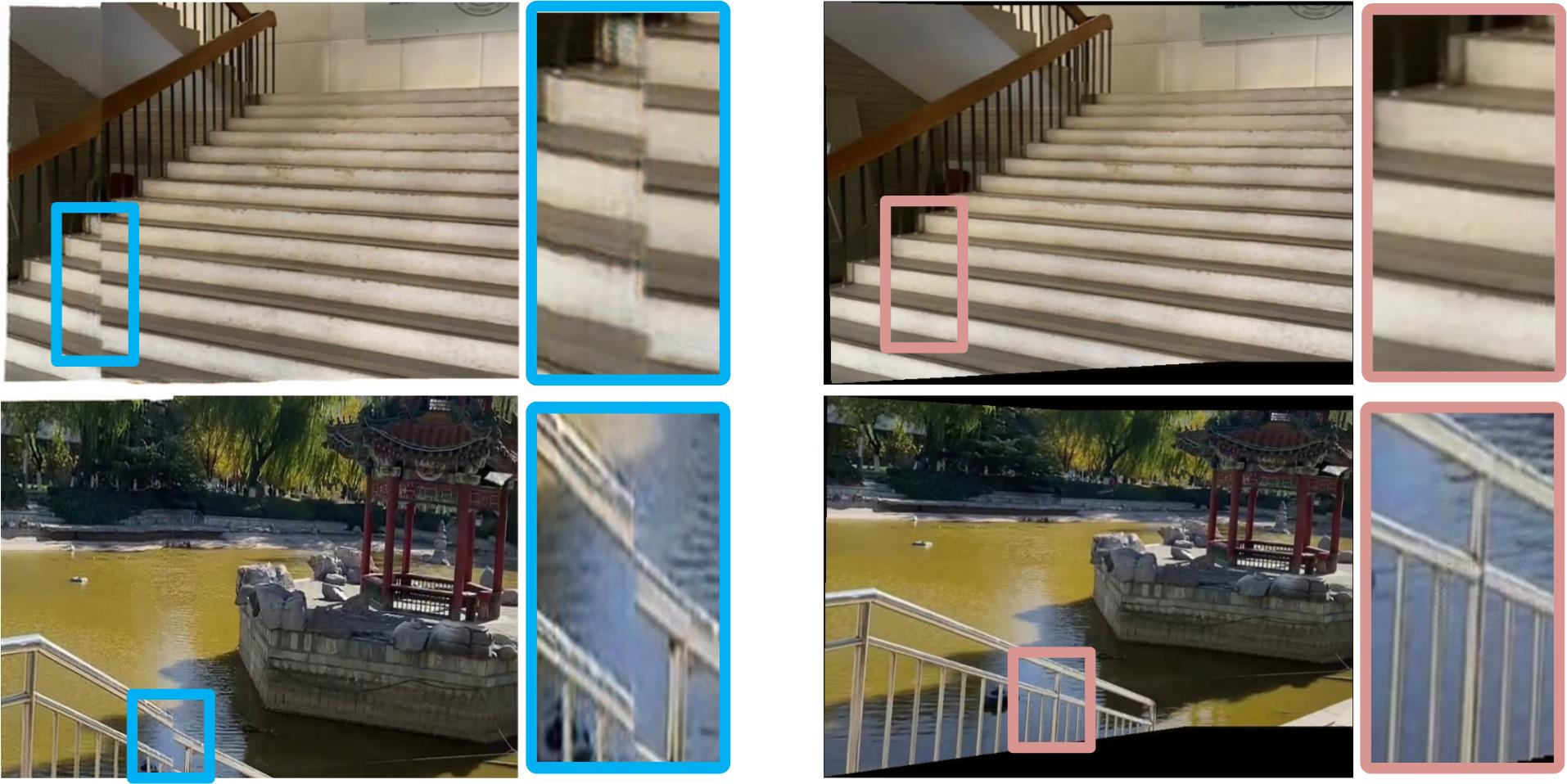}}
    \subfigure[Comparison of illumination difference. Left: EPISNet\cite{nie2020learning}. Right: ours.]
    {\includegraphics[width=0.45\textwidth]{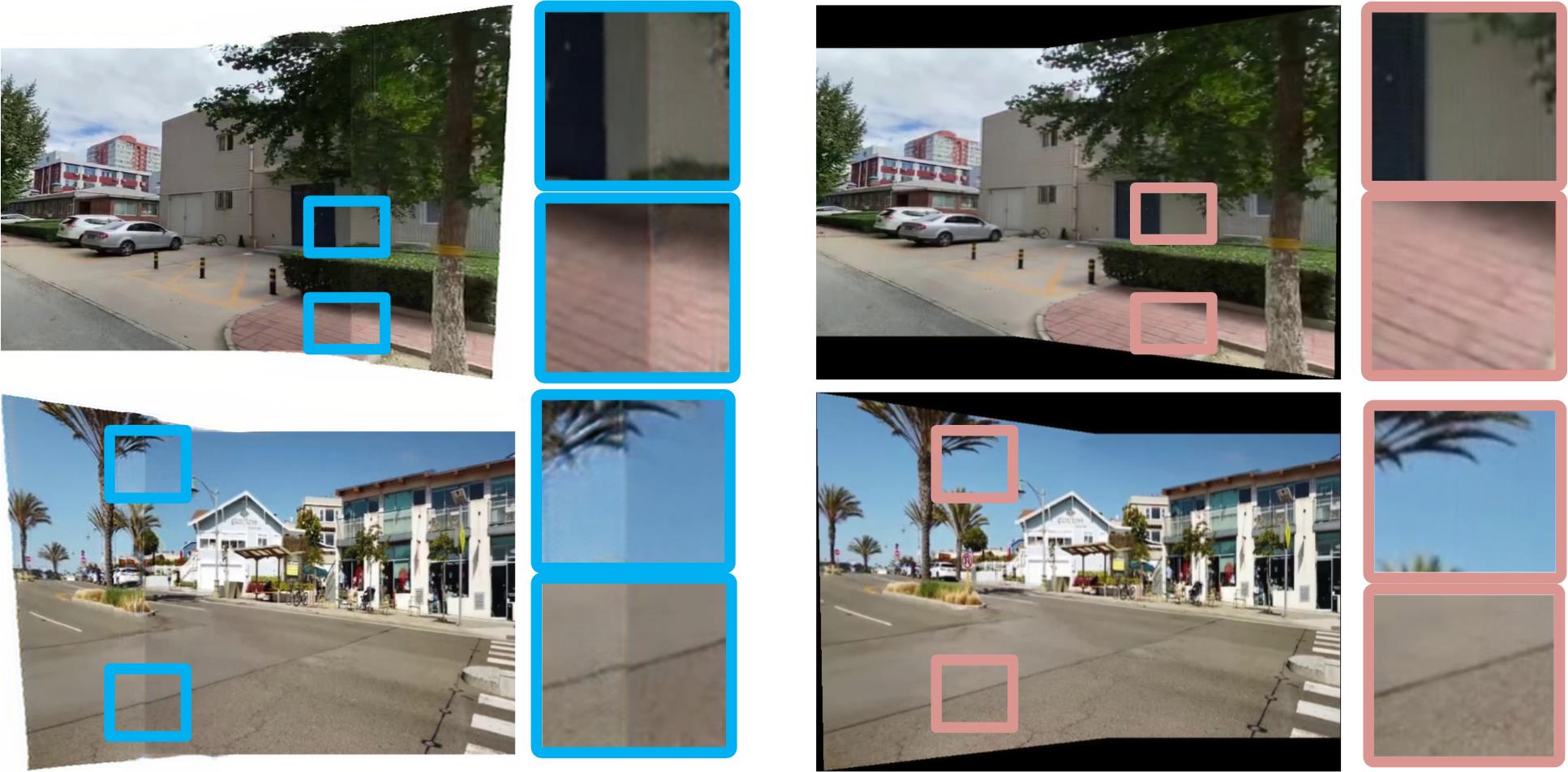}}
    \vspace{-0.2cm}
    \caption{Study on continuity and illumination.}
    %\label{fig:long}
    %\label{fig:onerow}
    \label{compare_2}
 \end{figure}

\begin{spacing}{1.5}
\end{spacing}
\textbf{Study on Continuity.}
The supervised deep image stitching methods \cite{nie2020view, nie2020learning} sacrifice the continuity of the edges (the edges between the reference image and the non-overlapping areas of the target image) to minimize artifacts. Although an edge-preserved network is proposed in EPISNet to weaken this problem, this problem still exists in a few testing cases. The discontinuity is demonstrated in the left picture of Fig. \ref{compare_2} (a), where discontinuous areas are framed and enlarged. This problem is settled perfectly in our unsupervised approach, as shown in the right picture of Fig. \ref{compare_2} (a). It gives credit to our constraint on seam masks, which enforces the edges of the overlapping areas close to one of the warped images.

\begin{spacing}{1.5}
\end{spacing}
\textbf{Study on Illumination.}
Another advantage of our method is that ours can smooth the illumination difference between the two images. The comparison with EPISNet are illustrated in \ref{compare_2} (b). The supervised methods fail to smooth the illumination difference because they are trained in a synthetic dataset with no illumination difference in the input images (the supervised methods cannot be trained in a real dataset due to the lack of stitched labels). On the contrary, our method is trained in real scenes, which can effectively learn how to smooth the illumination difference caused by different shooting positions.

\begin{spacing}{1.5}
\end{spacing}
\textbf{Study on Visual Quality.}
\begin{figure}[!t]
    \centering
    {\includegraphics[width=0.47\textwidth]{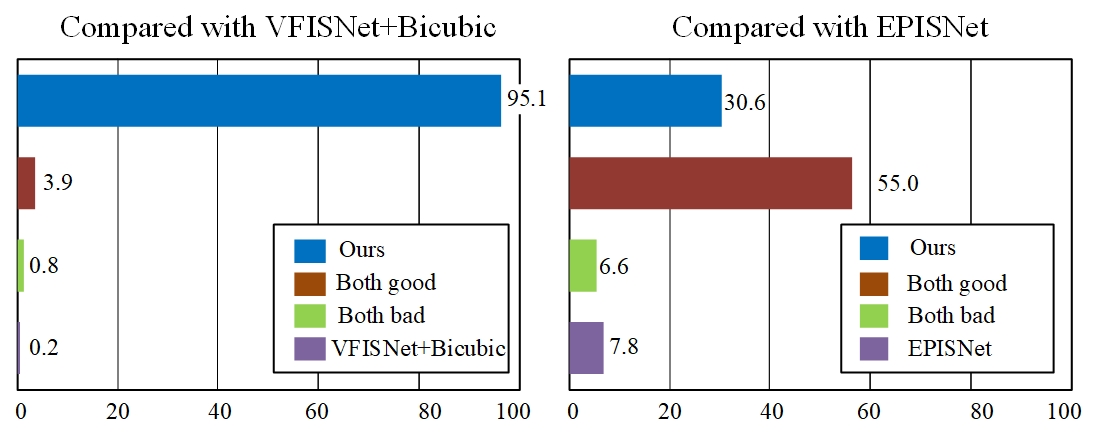}}
    \vspace{-0.4cm}
    \caption{User study on visual quality: compared with learning-based methods. The numbers are shown in percentage and averaged on 20 participants.}
    %\label{fig:long}
    %\label{fig:onerow}
    \label{user_2}
 \end{figure}
Similar to the user study with feature-based methods, we adopt the same strategy to investigate every participant to compare our method with the existing learning-based ones. Considering VFISNet can only work on the resolution of $128\times 128$, we use Bicubic interpolation to resize the stitched images. The results are shown in Fig. \ref{user_2}. Since Bicubic interpolation inevitably brings blurs when zooming in on images, the probability of preferring our method is further greater than that of preferring VFISNet+Bicubic. Even compared with EPISNet, our method is still preferred on the visual quality of the stitched images.

Besides that, Fig. \ref{real_results} exhibits the visual comparative results with these supervised methods, where the green rectangles indicate the severely blurred regions and the red rectangles point to discontinuous edges.

\begin{figure*}[!t]
    \centering
    \subfigure[Results on classic instances outside of our dataset. From left to right: ``roof"\cite{zhang2020content}, ``theater"\cite{li2019local}, ``street"\cite{he2016parallax}, ``roadside"\cite{chang2014shape}, and ``officedesk"\cite{li2019local}.]
    {\includegraphics[width=1\textwidth]{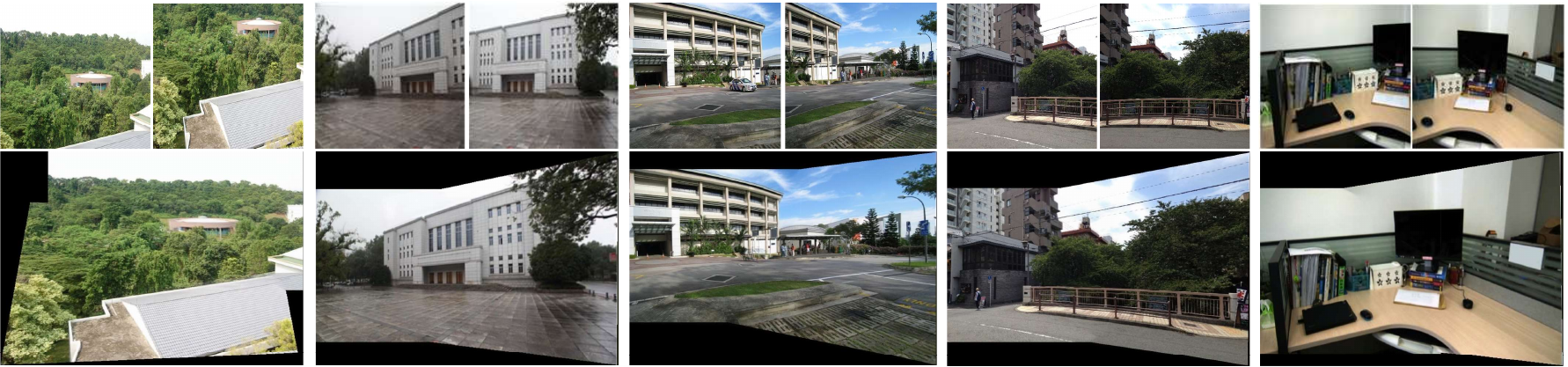}}
    \subfigure[Results on our proposed dataset. From left to right: ``stairs", ``snow", ``grass", ``lake", and ``campus".]
    {\includegraphics[width=1\textwidth]{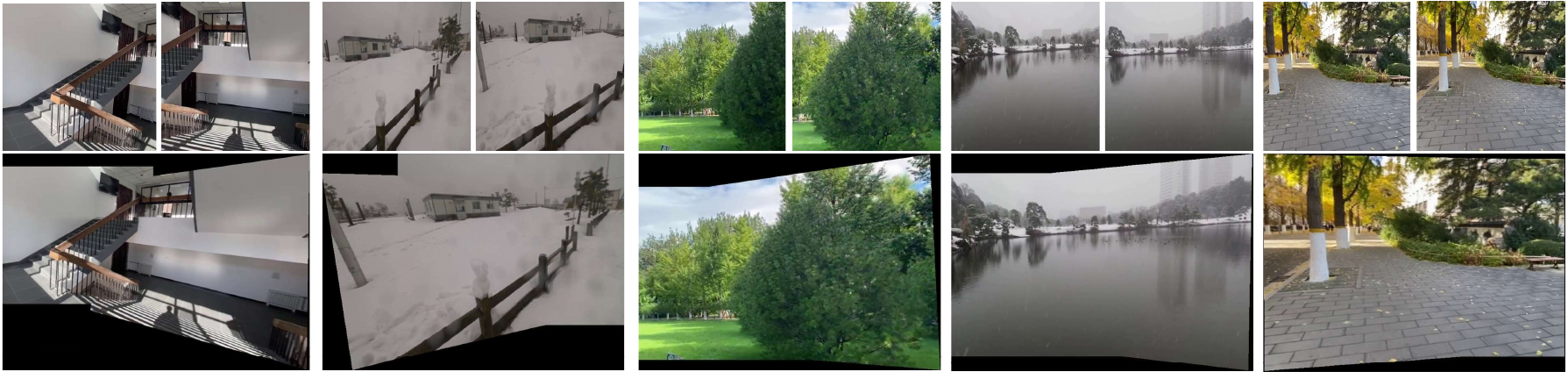}}
    \vspace{-0.2cm}
    \caption{More results of ours.}
    %\label{fig:long}
    %\label{fig:onerow}
    \label{more_results}
 \end{figure*}

\begin{spacing}{1.5}
\end{spacing}
To perceive our visual quality more intuitively, more results are illustrated in Fig. \ref{more_results}, where the inputs and the outputs are demonstrated together.

\subsection{Ablation Studies}
\label{section54}

\begin{table}[]
    \centering
   \caption{Frameworks for ablation studies.}
   \scalebox{0.77}{
    \begin{tabular}{|l||l|l||l|l|l|}
        \hline
        \multirow{2}{*}{}   & \multicolumn{2}{c||}{Architecture} & \multicolumn{3}{c|}{Loss} \\
        \cline{2-6}
        & LR branch  & HR branch & Content loss & Seam loss & CS loss \\
 \hline
 v1 & \makecell[c]{\checkmark}  &  & \makecell[c]{\checkmark} &  &  \\
 \hline
 v2 & \makecell[c]{\checkmark}  & \makecell[c]{\checkmark} & \makecell[c]{\checkmark} &  &  \\
 \hline
 v3 & \makecell[c]{\checkmark}  & \makecell[c]{\checkmark} & \makecell[c]{\checkmark} & \makecell[c]{\checkmark} &  \\
 \hline
 Ours & \makecell[c]{\checkmark}  & \makecell[c]{\checkmark} & \makecell[c]{\checkmark} & \makecell[c]{\checkmark} & \makecell[c]{\checkmark} \\
 \hline
    \end{tabular}
   }
    \label{ablation_table}
\end{table}

\begin{figure}[!t]
    \centering
    {\includegraphics[width=0.46\textwidth]{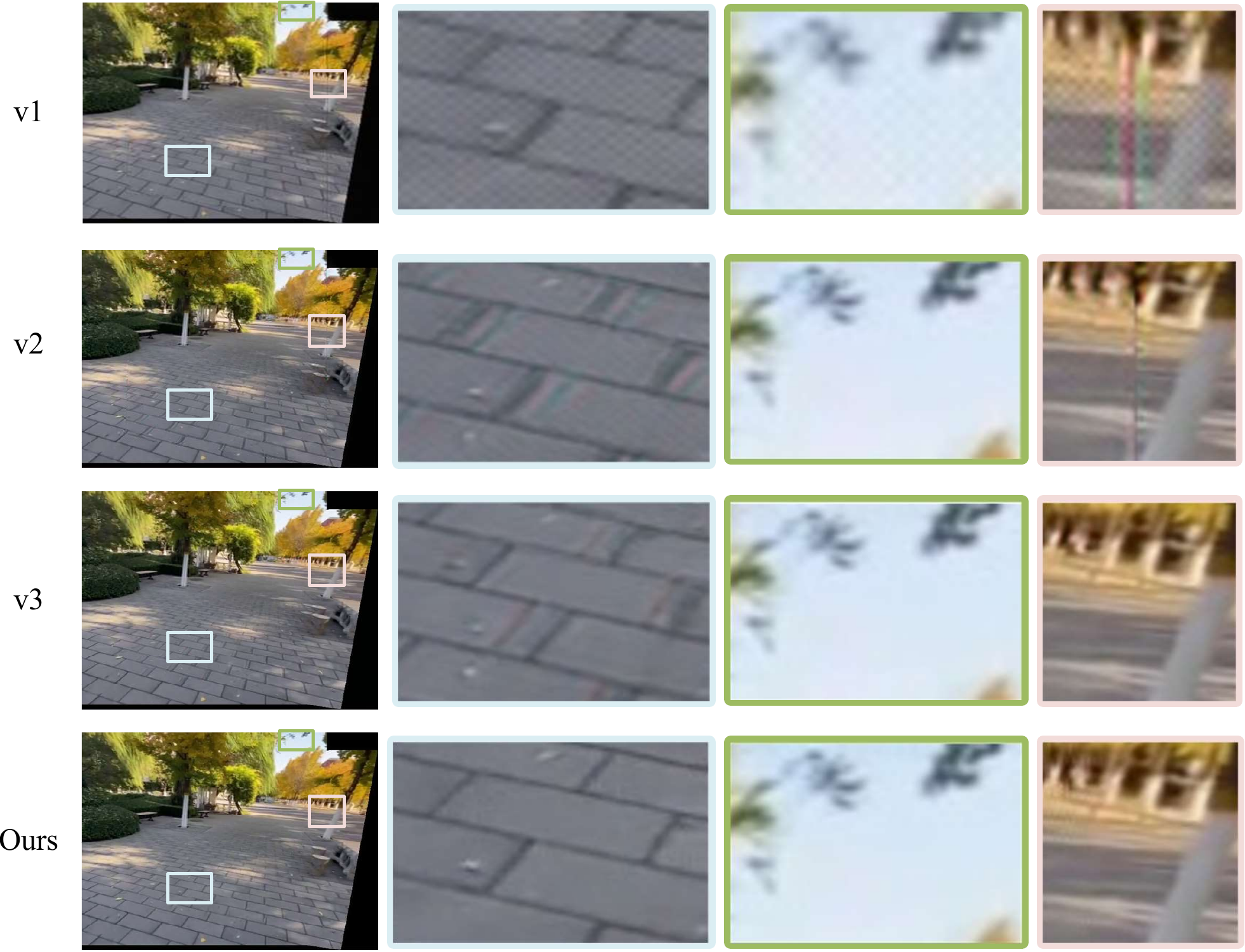}}
    \vspace{-0.3cm}
    \caption{Ablation studies on our framework. Col 1: outputs of different frameworks. Col 2-4: enlarged image patches to show the differences on artifacts, definition, and seam distortions, respectively.}
    %\label{fig:long}
    %\label{fig:onerow}
    \label{ablation1}
 \end{figure}

In this section, ablation studies are performed on both network architectures and loss functions. In the architecture, we validate the effectiveness of the low-resolution branch (LR branch) and high-resolution branch (HR branch); in the loss, we test the function of the content loss, seam loss, and content consistency loss (CS loss). The properties of all the studied frameworks are shown in Table \ref{ablation_table}.

From the results which are illustrated in Fig. \ref{ablation1}, we can observe:

(1) The most straightforward combination of LR branch and content loss can realize image stitching. However, there are still two issues unresolved: seam distortions (row 1, col 4 in Fig. \ref{ablation1}) and limited resolution. In our analysis, the seam distortion is the side effect of the proposed content loss.

(2) Compared v2 with v1, the HR branch can effectively enhance the resolution of the stitched image. As the cost, a few artifacts (row 2, col 2 in Fig. \ref{ablation1}) are introduced since the receptive field of HR branch convolution kernels is too small for higher resolution images.

(3) Compared with v2, v3 removes the seam distortions (row 3, col 4 in Fig. \ref{ablation1}) using the proposed seam loss. By imposing a pixel-level similarity constraint on the edge of the overlapping area, the seam distortions are suppressed successfully. However, there are still artifacts (row 3, col 2 in Fig. \ref{ablation1}) in the stitched image.

(4) Compared with v3, ours removes the artifacts (row 4, col 2 in Fig. \ref{ablation1}) using the proposed CS loss. The CS loss serves as an enhancer of the receptive field, which promotes the receptive field of the HR branch from that of the LR branch.

\begin{figure}[!t]
    \centering
    {\includegraphics[width=0.47\textwidth]{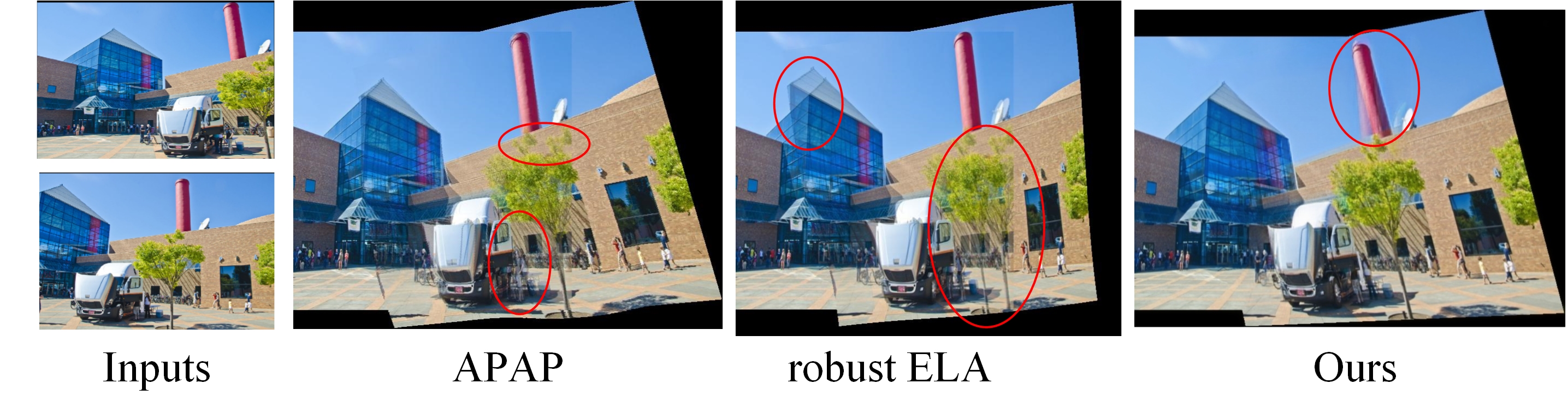}}
    \vspace{-0.4cm}
    \caption{A failure example. The red circle indicates the unsatisfying stitched areas.}
    %\label{fig:long}
    %\label{fig:onerow}
    \label{limitaion}
 \end{figure}

%------------------------------------------------------------------------
\section{Limitation and Future Work}
\label{section6}
The proposed solution eliminates parallax artifacts through reconstructing the stitched images from feature to pixel. It is still essentially a stitching method based on a single homography. As the parallax increases, the alignment performance of the first stage will decrease, while the burden of the reconstruction network will also become heavier. When the parallax is too large, the reconstruction network may treat the misalignments as new objects to reconstruct. An example is shown in Fig. \ref{limitaion}. In the future, we hope to solve this problem in two directions: 1) Improve the alignment performance of the alignment network to decrease the burden of the reconstruction network. 2) Increase the receptive field of the reconstruction network to deal with remained large misalignments.

%------------------------------------------------------------------------
\section{Conclusion}
\label{section7}
This paper proposes an unsupervised deep image stitching framework, comprising unsupervised coarse image alignment and unsupervised image reconstruction. In the alignment stage, an ablation-based loss function is proposed to constrain the unsupervised deep homography estimation in large-baseline scenes, and a stitching-domain transformer layer is designed to warp the input images in the stitching-domain space. In the reconstruction stage, an unsupervised deep image reconstruction network is proposed to reconstruct the stitched images from feature to pixel, eliminating the artifacts in an unsupervised reconstruction manner. Besides, a real dataset for unsupervised deep image stitching is presented, which we hope can work as a benchmark dataset for other methods. Experimental results demonstrate the superiority of our method over other state-of-the-art solutions. Even if compared with the supervised deep image stitching solutions, the results of our unsupervised approach are still preferred by users in terms of visual quality.

However, the reconstruction ability is not unlimited, which indicates our solution may fail in the scenes with extremely large parallax. Considering our first stage is essentially an alignment model based on a single homography, the ability to handle large parallax can be improved by extending the linear deep homography network to a non-linear homography model. Moreover, the reconstruction performance can be further increased by increasing the receptive field of the reconstruction network, which is also an exploring direction of the future work.

\normalem
\bibliographystyle{ieeetr}
\bibliography{reference}

\end{document}